%% file: main.tex
\documentclass[letterpaper]{article} 
\usepackage[preprint]{aaai2027}    

\usepackage[hyphens]{url}            
\usepackage{graphicx}                
\usepackage{natbib}                  
\usepackage{caption}                 
\usepackage{algorithm}
\usepackage{algorithmic}

\usepackage{amsmath}
\usepackage{amssymb}

\usepackage{booktabs}
\usepackage{array}
\usepackage{adjustbox}
\usepackage{multirow}
\usepackage{enumitem}
\usepackage{amsthm}
\newtheorem{remark}{Remark}

\title{Reward-Guided Fine-Tuning of One-Step Generative Models \\
  via Wasserstein Gradient Flow}

\author{
Hoseong Hwang\textsuperscript{\rm 1}\equalcontrib, 
Woorim Han\textsuperscript{\rm 1}\equalcontrib,
Joungin Chun\textsuperscript{\rm 1},
Jinseong Park\textsuperscript{\rm 2}\corresponding,
Jaewoong Choi\textsuperscript{\rm 1}\corresponding
}

\affiliations{
    \textsuperscript{\rm 1}Sungkyunkwan University, Seoul, Republic of Korea \\
    \textsuperscript{\rm 2}Korea Institute for Advanced Study, Seoul, Republic of Korea \\
    \{hhs333, woorim, juchun2003\}@skku.edu, jinseong@kias.re.kr, 
    jaewoongchoi@skku.edu
}

\begin{document}

\maketitle

\begin{abstract}
To mitigate the time complexity of generative models, one-step generative models have recently emerged through direct mapping from noise to data in a single forward pass. However, the reward-guided fine-tuning method of one-step generative models remains largely unexplored. To address this, we consider one-step generators from an optimal transport view, investigating Wasserstein Gradient Flow (WGF) for modeling smooth and controlled distributional evolution in probability space. We then propose a novel reward-guided fine-tuning of a one-step generative model via WGF. We derive a practical training method that requires no reward gradients, thereby handling both non-differentiable and differentiable rewards. Moreover, our method provides smooth and stable reward-guided distributional updates while mitigating reward hacking and mode collapse. Experiments on 2D synthetic data, CIFAR-10, and ImageNet 256$\times$256 with diverse rewards, including JPEG (in)compressibility, class probability, Black-and-White and CLIP alignment, show that our method achieves better reward alignment compared to baselines.
\end{abstract}

\section{Introduction}
Diffusion models \citep{song2021scorebased, ho2020denoising} have emerged as the leading paradigm for data generation, achieving high-quality image synthesis. However, their widespread use is heavily bottlenecked by slow inference speeds due to their iterative denoising steps. To address this computational limitation, recent research has focused on one-step generative frameworks, such as consistency models and flow maps \citep{kim2023consistency, geng2025mean, frans2025one}. By directly mapping from the noise distribution to the data distribution in a single forward pass, these models achieve efficient generation while preserving sample quality.

Recent advances in fine-tuning have demonstrated remarkable success across diverse tasks, including image aesthetics \cite{black2023training, clark2024directly, prabhudesai2023aligning}, image compression \cite{fan2023dpok, black2023training}, and text-image alignment \cite{black2023training, clark2024directly}. An especially promising direction is performing reward fine-tuning in the one- or few-step frameworks, where the model generates samples in one or a few forward passes to reduce computational cost. Consequently, recent studies \citep{shekhar2025rocm, jia2025reward, oertell2024rl} have increasingly highlighted the importance of fine-tuning in the one- or few-step frameworks.

Despite this growing interest in fast inference, existing fine-tuning methods designed for multi-step diffusion models \cite{black2023training, fan2023dpok} cannot be directly applied to one-step generation. These methods operate on the intermediate states of an iterative denoising trajectory, which a one-step generator does not produce. Moreover, one-step models generally follow a linear generation path \cite{uotm}, unlike the non-linear trajectory assumed by diffusion-based fine-tuning.

To address this gap, we propose a novel fine-tuning method for one-step generative models via \textbf{Wasserstein Gradient Flow (WGF)}. 
WGF provides a suitable framework for fine-tuning one-step generative models via smooth and controlled distributional dynamics in Wasserstein space \cite{ambrosio}. In contrast to diffusion-based fine-tuning frameworks that rely on iterative multi-step trajectories, WGF directly optimizes the evolution of the model distribution in probability space. As a result, fine-tuning via WGF enables stable distributional updates without requiring iterative inference. Moreover, by operating in Wasserstein space, the model evolves through conservative modifications around the pre-trained distribution, effectively mitigating reward hacking and mode collapse.

We formulate reward-guided fine-tuning as a WGF problem that evolves the pre-trained distribution toward a reward-weighted target. To make this tractable, we discretize the continuous WGF using the JKO scheme and derive a practical adversarial learning objective via a semi-dual formulation \cite{s-jko} and importance sampling. Our contributions are summarized as follows:
\begin{itemize}
    \item We present the first framework for reward-guided fine-tuning of one-step generative models based on the mathematical foundations of WGF and optimal transport.
    \item We derive a practical training algorithm that handles both non-differentiable and differentiable rewards via the JKO scheme and importance sampling.
    \item Experiments on 2D synthetic data, CIFAR-10, and ImageNet 256$\times$256 demonstrate that our method achieves a better reward alignment than baselines. 
\end{itemize}

\section{Preliminaries}

\paragraph{Notations and Assumptions.}
Let $\mathcal{P}(\mathbb{R}^d)$ be the set of probability distributions on $\mathbb{R}^d$ that are absolutely continuous with respect to the Lebesgue measure. For a measurable map $T$, $T_\# \mu$ represents the pushforward distribution of $\mu$. 
The 2-Wasserstein distance $\mathcal{W}_2 (\cdot, \cdot)$ is defined as follows:
\begin{equation} \label{eq:2-was}
    \mathcal{W}_2 (\rho, \xi) := \left(\min_{\pi \in \Pi(\rho, \xi)} \int_{\mathbb{R}^d \times \mathbb{R}^d} \lVert x-y \rVert_2^2 d\pi(x,y)\right)^{\frac{1}{2}},
\end{equation}
where $\Pi(\rho, \xi)$ denotes the set of joint probability distributions on $\mathbb{R}^d\times\mathbb{R}^d$ whose marginals are $\rho$ and $\xi$. 
Moreover, $f^*$ indicates the convex conjugate of a function $f$, i.e., $f^{*}(y) = \sup_{x \in \mathbb{R}}\{\langle x, y \rangle - f(x)\}$ for $f:\mathbb{R}\rightarrow [-\infty, \infty]$.

\subsection{One-Step Generative Models}

Diffusion models \citep{song2021scorebased} and flow matching \citep{lipman2023flow,liu2022flow} learn a continuous transformation from a tractable noise distribution $p_0$ to the target data distribution $p_1 = p_{\text{data}}$.
This continuous transformation is modeled as a probability flow via an Ordinary Differential Equation (ODE) as follows: 
\begin{equation} \label{eq:pf_ode}
\mathrm{d}x_t = v_\theta(x_t, t) \mathrm{d}t, \quad t \in [0, 1],
\end{equation}
where $v_{\theta}$ denotes a velocity field over a continuous time $t \in [0,1]$. In diffusion models, $v_{\theta}$ is implicitly defined via the score function \citep{song2021scorebased}, while in flow matching it is trained through a direct regression of ground-truth conditional vector field \citep{lipman2023flow,liu2022flow}. To generate a new sample, the probability flow ODE (PF-ODE) defined by $v_{\theta}$ is solved from $t =0$ to $t =1$, starting with an initial value $x_0 \sim \mathcal{N}(0,I)$. 

Since solving this ODE over $t \in [0,1]$ requires numerical integration, the resulting iterative procedure involves tens to hundreds of neural network evaluations, leading to computationally expensive inference. To enable efficient one- or few-step generation by directly learning the solution to the probability flow (Eq. \ref{eq:pf_ode}), recent models such as Consistency Trajectory Model (CTM) \cite{kim2023consistency} and MeanFlow \cite{geng2025mean} distill or parameterize the flow map, which maps a state at time $t$ directly to time $s$:
\begin{equation}
    \psi(x_t, t, s) = x_t + \int_{t}^{s} v_\theta(x_\tau, \tau) \mathrm{d}\tau.
\end{equation}
By directly learning this mapping, the entire generation process can be executed in a single forward pass: 
\begin{equation}
    x_1 = \psi(x_0, 0, 1) = G_\theta(x_0).
\end{equation}
In this work, we assume access to a pre-trained one-step generative model $T_0 : \mathbb{R}^d \to \mathbb{R}^d$ such that $(T_0)_\#\mu \approx \nu$, where $\mu$ is a Gaussian prior and $\nu$ is the data distribution.

\subsection{Wasserstein Gradient Flow and JKO scheme}

\paragraph{Wasserstein Gradient Flow.} 
Given a functional $\mathcal{F}(\rho)$ on $\rho\in \mathcal{P}(\mathbb{R}^d)$, the Wasserstein Gradient Flow (WGF) \cite{ambrosio} describes the dynamics of a probability density $\{\rho_{t}\}_{t \geq 0}$, following the steepest descent direction of $\mathcal{F}(\rho)$. Here, the metric on $\mathcal{P}(\mathbb{R}^d)$ is defined as the 2-Wasserstein distance $\mathcal{W}_2$ (Eq. \ref{eq:2-was}). 
The WGF can be explicitly written as the PDE as follows:
\begin{equation} \label{eq:WGF}
    \frac{\partial \rho}{\partial t} = \nabla \cdot \left( \rho \nabla \frac{\delta \mathcal{F}}{\delta \rho} \right),
\end{equation}
where $\frac{\delta\mathcal{F}}{\delta \rho} $ denotes the first variation of $\mathcal{F}$ with respect to the standard $L_2$ metric \cite{villani}.

When $\mathcal{F}(\rho)$ is given as the $f$-divergence $D_f$ with respect to the target distribution $\nu$, \textbf{WGF describes the trajectories of probability density $\{\rho_{t}\}_{t \geq 0}$ evolving from $\mu=\rho_{0}$ towards $\nu$} by minimizing $\mathcal{F}(\rho)$:
\begin{equation} \label{eq:f-div}
    \mathcal{F}(\rho) := D_f (\rho | \nu) = \int f \left( \frac{d \rho}{d \nu} \right) d \nu.
\end{equation}
$\rho_0$ corresponds to the pre-trained model distribution and $\nu$ corresponds to the reward-weighted distribution $\nu_r$.

\paragraph{JKO Scheme.}
Computing the continuous WGF is a challenging problem. To address this, \citet{jko} proposed the \textbf{JKO scheme}, which uses a time discretization scheme to approximate the WGF. In this scheme, given the current JKO step $\mu_{k}$, the next JKO step $\mu_{k+1}$ is formally defined as follows:
\begin{equation} \label{eq:jko}
    \mu_{k+1} = \underset{\rho \in \mathcal{P}(\mathbb{R}^d)}{\text{argmin}} \left[ \frac{1}{2h} \mathcal{W}_2^2 (\rho, \, \mu_{k}) + \mathcal{F}(\rho) \right],
\end{equation}
where $\mu_{0}=\mu$ is the initial condition. Intuitively, $h$ can be understood as the step size of the time discretization. As $h \to 0$, the sequence $\{\mu_k\}$ converges to the continuous WGF trajectory $\{\rho_{kh}\}$.

\section{Method}
In this section, we propose a new reward-guided fine-tuning of one-step generative models via WGF. In Section~\ref{sec:formulation}, we formulate reward-guided fine-tuning as a WGF problem, describing the dynamics from the pre-trained distribution $\nu$ to the reward-weighted target $\nu_r$. In Section~\ref{sec:jko_reformulation}, we discretize these continuous dynamics by the JKO scheme and reformulate each JKO step via its semi-dual form, presenting its advantage in the following Section~\ref{sec:adv_wgf}.

\subsection{Problem Formulation} \label{sec:formulation}
Our goal is to fine-tune a one-step generative model $T_0$ to generate samples from a reward-weighted distribution $\nu_{r}$. Let $T_0: \mathbb{R}^{d} \to \mathbb{R}^{d}$ be a pre-trained one-step generative model trained on an uncontrolled data distribution $\nu$ with a (Gaussian) prior distribution $\mu = \mathcal{N}(0, I)$, such that $(T_0)_{\#} \mu \approx \nu$. Here, the reward-weighted distribution is defined as 
\begin{multline} \label{eq:reward_weighted_target}
    \nu_{r} (y) = \frac{\exp \left(\beta \cdot r(y) \right)}{M_{\beta}} \nu(y) \\
    \text{where} \quad M_{\beta} = \int \exp( \beta \cdot r(y) ) d\nu(y),
\end{multline}
$M_\beta$ denotes the normalizing constant and $\beta >0$ is a reward intensity parameter controlling the trade-off between the data distribution and the reward. Here, $r: \mathbb{R}^{d} \to \mathbb{R}$ is a given reward function, where we can consider both differentiable and non-differentiable rewards, such as compression ratio, or classifier output. 
Moreover, we assume that we do not have direct access to samples from $\nu_r$. 

We formulate fine-tuning as minimizing the $f$-divergence between the model distribution and the reward-weighted target via WGF. Specifically, we define the energy functional:
\begin{equation} \label{eq:wgf_finetuning}
\mathcal{F}_r (\rho) = D_f(\rho \mid \nu_r) \quad \text{for } \rho \in \mathcal{P}(\mathbb{R}^d),
\end{equation}
which measures the divergence between the current distribution and the reward-weighted target $\nu_r$. The WGF $\{ \rho_{t} \}_{t \geq 0}$ starting from $\rho_0 = (T_0)_\#\mu$ describes the evolution of the probability density from the distribution of the pre-trained model $\nu$ to $\nu_r$ by following the steepest descent of $\mathcal{F}_r(\rho)$ in the Wasserstein space $\mathcal{W}_2$,
\begin{equation} \label{eq:fine-tuning_pde}
\frac{\partial \rho}{\partial t} = \nabla \cdot \left(\rho \nabla \frac{\delta \mathcal{F}_r}{\delta \rho}\right), \quad \rho_0 = (T_0)_\#\mu.
\end{equation} 

\subsection{Fine-tuning via JKO Scheme} \label{sec:jko_reformulation}
Computing the continuous WGF in Eq. \ref{eq:fine-tuning_pde} is a challenging problem. To address this, we leverage the JKO scheme \citep{jko} for fine-tuning. Under the JKO scheme, we conduct the following iterative update:
\begin{equation} \label{eq:jko_finetuning}
\begin{aligned}
\nu_{k+1} 
&= \underset{\rho \in \mathcal{P}(\mathbb{R}^d)}{\mathrm{argmin}} 
\; \underbrace{
\frac{1}{2h} \mathcal{W}_2^2 (\nu_{k}, \rho) 
+ D_f (\rho \mid \nu_{r})
}_{\mathcal{L}_{\rho}} \\
&\quad \text{with} \quad 
\nu_{0} = T^{\theta}_{\#} \mu \approx \nu.
\end{aligned}
\end{equation}
where $h>0$ indicates the step size in the time discretization. The JKO scheme balances two objectives: the Wasserstein distance term $W_2^2(\nu_k, \rho)$ ensures that the updated distribution stays close to the previous step $\nu_{k}$, while the divergence functional $D_f (\rho \mid \nu_{r})$ leads the distribution toward the reward-weighted target. As the step size $h \to 0$, the JKO step $\{\nu_k\}$ converges to the continuous WGF trajectory $\{\rho_{kh}\}$.

\paragraph{Semi-dual Form of JKO.}
Following \citet{s-jko}, the JKO step admits a semi-dual formulation. Importantly, this formulation allows us to directly parametrize each JKO step using a transport map and a potential function $v_{\phi}$, without explicitly modeling the probability density. By leveraging the equivalence with the Unbalanced Optimal Transport (UOT) problem \citep{uot1, uot2}, this semi-dual formulation is given as follows:
\begin{equation} \label{eq:semi-dual-jko}
    \sup_{v \in \mathcal{C}} \int v^c(x) d\nu_k(x) + \int f^\circ(v(y)) d\nu_{r}(y),
\end{equation}
where $v: \mathbb{R}^d \to \mathbb{R}$ is a potential function, $v^c(x) := \inf_y ( c_h(x,y) - v(y) )$ is its $c$-transform with $c_h(x,y) = \frac{1}{2h}\|x-y\|_2^2$, and $f^\circ(x) := -f^*(-x)$ with $f^*$ being the convex conjugate of $f$. Here, we implicitly represent the evolving transport map from the $k$-th JKO step $\nu_{k}$ to the $(k+1)$-th JKO step $\nu_{k+1}$ via $c$-transform parameterization \citep{otm, fanTMLR}:
\begin{equation}
    \Delta T_{k} : x \mapsto \underset{y}{\arg\min} \left( c_h(x,y) - v(y) \right).
\end{equation}
Then, $\Delta T_{k}$ satisfies the following:
\begin{equation}
    v^c(x) := c_h(x,\Delta T_{k}(x)) - v(\Delta T_{k}(x)).
\end{equation} 
Substituting this into Eq.~\eqref{eq:semi-dual-jko} yields the max-min objective for a single JKO step as follows:
\begin{equation} \label{eq:semi-dual-jko-with-transport}
\begin{aligned}
\sup_{v \in \mathcal{C}} \; & \int \inf_{\Delta T_{k}} 
\left[ 
c_h\left(x, \Delta T_{k}(x)\right) - v(\Delta T_{k}(x)) 
\right] d\nu_k(x) \\
&+ \int f^\circ(v(y)) d\nu_{r}(y).
\end{aligned}
\end{equation}
The objective for a single JKO step is equivalent to the
source-fixed UOTM between $\nu_k$ and $\nu_{k+1}=(\Delta T_k)_\#\nu_k$
\citep{s-jko}, which allows our fine-tuning method to be naturally interpreted as solving a sequence of optimal transport problems. Additionally, from an adversarial learning perspective, we can interpret $\Delta T_k$ as a generator that maps $\nu_k$ to $\nu_{k+1}$, and the potential function $v$ is interpreted as a discriminator between $\nu_k$ and $\nu_{k+1}$.

\paragraph{Fine-tuning Parameterization.}
In Eq.~\ref{eq:semi-dual-jko-with-transport}, approximating the integral requires sampling from $\nu_k$, which entails simulating the entire trajectory $\nu_0 \to \cdots \to \nu_k$ of the JKO scheme. To avoid this simulation, we adopt a reparameterization, which replaces $\Delta T_k$-training with $T_k$-training and thus requires sampling only from the fixed initial distribution $\nu_0$. In our WGF formulation of fine-tuning, the initial distribution $\rho_{0} = \nu_0$ is the distribution of a pre-trained one-step generative model, and this WGF gradually converges to the reward-weighted target distribution $\nu_{r}$. In this respect, $\Delta T_k$ (Eq.~\ref{eq:semi-dual-jko-with-transport}) can be interpreted as a refinement that progressively transforms the uncontrolled distribution $\nu_0 \approx \nu$ (\textit{Source distribution}) into $\nu_r$. Let $T_k:\mathbb{R}^d \to \mathbb{R}^d$ denote the cumulative transformation after the $k$ refinement steps:
\begin{equation}
    T_k = \Delta T_{k}  \circ \cdots \circ \Delta T_{1} \circ \Delta T_{0}.
\end{equation}
Then, $T_k$ satisfies $T_{k} := \Delta T_{k} \circ T_{k-1}$ and $(T_k)_\# \mu = \nu_{k+1}$. Here, $T_k$ represents our \textbf{fine-tuned generative model} after the $k+1$-th JKO step. 
Substituting this reparameterization into Eq. \ref{eq:semi-dual-jko-with-transport} yields the following objective for the $k$-th step:
\begin{equation} \label{eq:finetuning_repara}
\begin{aligned}
\mathcal{L}_{k} 
= \sup_{v \in \mathcal{C}} \; 
& \int \inf_{T_{k}} 
\Big[
c_h\big(T_{k-1}(x), T_k(x)\big) - v(T_k(x))
\Big] d\mu(x) \\
&+ \int f^\circ(v(y)) d\nu_{r}(y).
\end{aligned}
\end{equation}

\paragraph{Importance Sampling for Reward-weighted Targets.}
The objective in Eq.~\ref{eq:finetuning_repara} requires samples from $\nu_{r}$, which are not directly accessible in our setting. We address this via importance sampling using the definition of $\nu_r$:
\begin{equation} \label{eq:importance_sampling_substitution}
\int f^\circ(v(y)) \, d\nu_{r}(y) 
= \int f^\circ(v(y)) \frac{\exp \left(\beta \cdot r(y) \right)}{M_{\beta}} \, d\nu(y).
\end{equation}
This allows us to estimate the integral using samples from $\nu$ (the pre-trained distribution), weighted by $\exp \left(\beta \cdot r(\cdot) \right)/M_{\beta}$. Substituting this into Eq. \ref{eq:finetuning_repara} yields the following objective for the $k$-th step:
\begin{equation} \label{eq:is}
\begin{aligned}
\mathcal{L}_{k} 
= \sup_{v_\phi \in \mathcal{C}} \;
& \int \inf_{T_\theta} 
\Big[
c_h\big(T_{k-1}(x), T_\theta(x)\big) - v_\phi(T_\theta(x))
\Big] d\mu(x) \\
&+ \int f^\circ(v_\phi(y)) \frac{\exp \left(\beta \cdot r(y) \right)}{M_{\beta}} \, d\nu(y),
\end{aligned}
\end{equation}
where we parameterize the potential function as $v_\phi$ and the transport map as $T_\theta$. The normalization constant $M_{\beta}$ is estimated using a moving average over mini-batches.

\paragraph{Pre-trained Distribution Updates.}
Eq.~\ref{eq:is} defines the reward-weighted term under the pre-trained distribution $\nu$. In practice, this fixed $\nu$ becomes increasingly mismatched with the current generator, leading to less stable optimization, consistent with the over-optimization behavior analyzed by \citet{gorbatovski2025learn}. Therefore, we update the pre-trained distribution $\nu$ during training $\nu \xleftarrow{\mathrm{sg}} \nu_k$, and use samples from the dynamically updated pre-trained distribution $\nu_k = (T_{k-1})_\# \mu$ to evaluate the reward-weighted term in Eq.~\ref{eq:is} (see Section~\ref{sec:ablation_study} for an ablation of the fixed and updated pre-trained distributions). This yields our final training objective:
\begin{equation} \label{eq:final_objective}
\begin{aligned}
\mathcal{L}_{k} 
= \sup_{v_\phi \in \mathcal{C}} \;
& \int \inf_{T_\theta} 
\Big[
c_h\big(T_{k-1}(x), T_\theta(x)\big) - v_\phi(T_\theta(x))
\Big] d\mu(x) \\
&+ \int f^\circ(v_\phi(y)) \frac{\exp \left(\beta \cdot r(y) \right)}{M_{\beta}} \, d\nu_{k}(y).
\end{aligned}
\end{equation}
The algorithm alternates between updating the generator (transport map $T_\theta$) and the discriminator (potential function $v_\phi$). We summarize the overall training algorithm in the Appendix~\ref{app:algorithm}.

\paragraph{Non-differentiable Rewards.}
Our method samples from the reward-weighted distribution via importance sampling, where the reward appears only as a scalar weight $\exp(\beta \cdot r(y))/M_{\beta}$ on $f^{\circ}(v_{\phi}(y))$. Since $y$ is drawn from the frozen model, our method does not require gradients through the reward function $r(\cdot)$. Therefore, our method can handle both differentiable and non-differentiable rewards.

\subsection{Advantages of WGF for Fine-tuning} \label{sec:adv_wgf}
\paragraph{Steepest Descent via Transport Maps.}
WGF provides a principled framework for fine-tuning by modeling the evolution of probability distributions along the steepest descent direction of the energy functional $\mathcal{F}_r (\rho) = D_f (\rho \mid \nu_{r})$. A key observation is that, through the semi-dual form of the JKO scheme, each steepest descent step can be realized directly as a transport map $T_{\theta}$ rather than as a continuous-time flow. The learned transport map moves samples along the steepest descent direction from the current distribution toward the target distribution, without learning a velocity field or solving an iterative ODE at inference time, in contrast to flow- or diffusion-based fine-tuning approaches. 

\begin{figure}[t]
    \centering

    \setlength{\fboxsep}{0pt}
    \setlength{\fboxrule}{0.1pt}
    \setlength{\tabcolsep}{0pt}

    \small

    \begin{tabular}{c @{\hspace{3mm}} c}
        \begin{tabular}{c}
            \small Pre-trained \\[1mm]
            \raisebox{-0.5\height}{\fbox{\includegraphics[width=0.20\columnwidth]{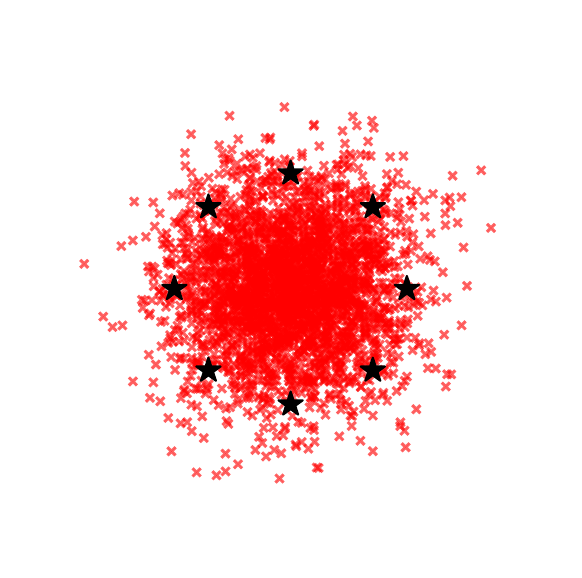}}} \\
            \\[-2mm] 
        \end{tabular}
        &
        \begin{tabular}{ccc @{\hspace{2mm}} c}
            \raisebox{-0.5\height}{\fbox{\includegraphics[width=0.20\columnwidth]{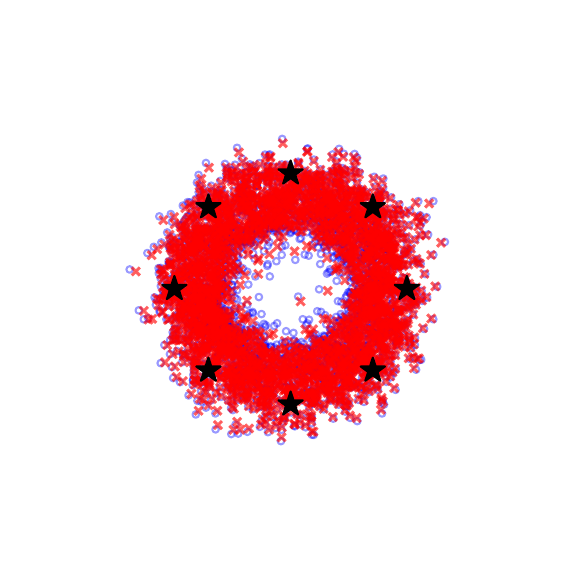}}} &
            \raisebox{-0.5\height}{\fbox{\includegraphics[width=0.20\columnwidth]{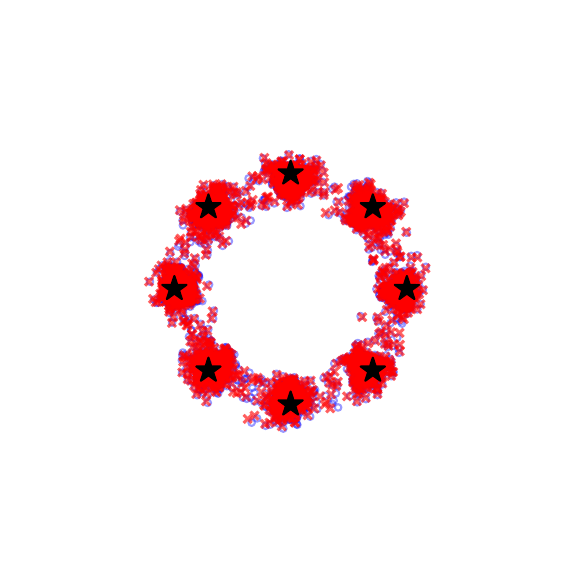}}} &
            \raisebox{-0.5\height}{\fbox{\includegraphics[width=0.20\columnwidth]{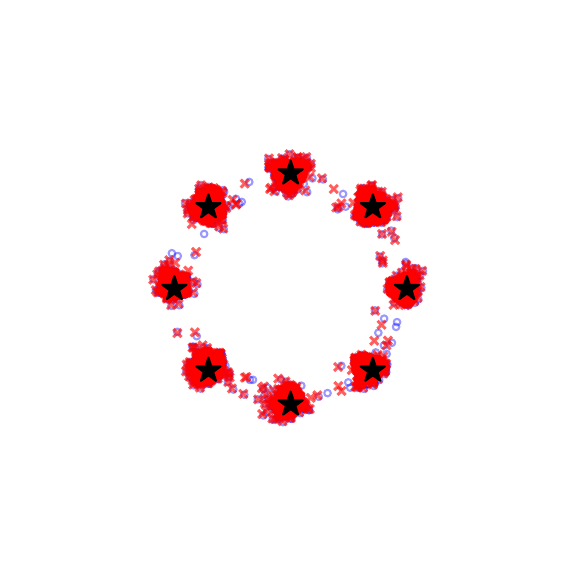}}} & 
            \raisebox{3mm}{\rotatebox{270}{\small Ours}} \\
            \small 5k & \small 15k & \small 25k & \\ [2mm]

            \raisebox{-0.5\height}{\fbox{\includegraphics[width=0.20\columnwidth]{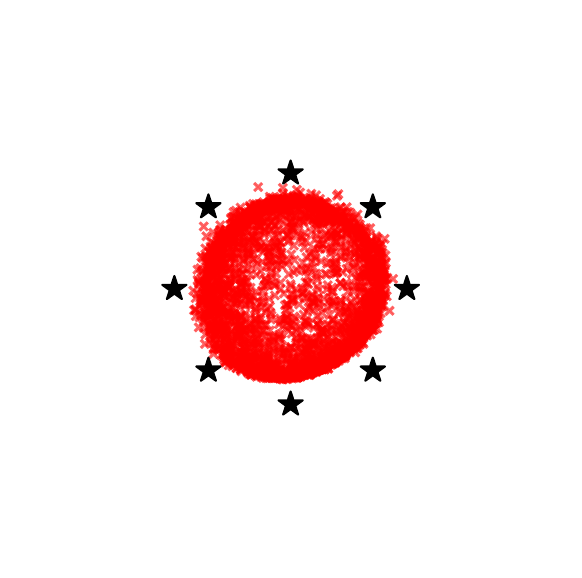}}} &
            \raisebox{-0.5\height}{\fbox{\includegraphics[width=0.20\columnwidth]{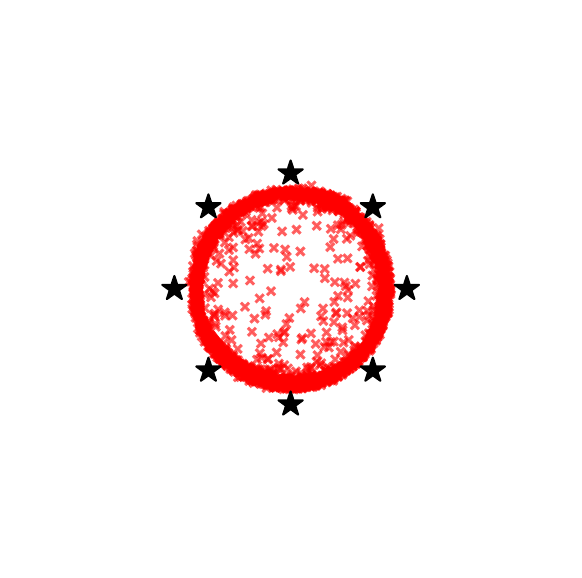}}} &
            \raisebox{-0.5\height}{\fbox{\includegraphics[width=0.20\columnwidth]{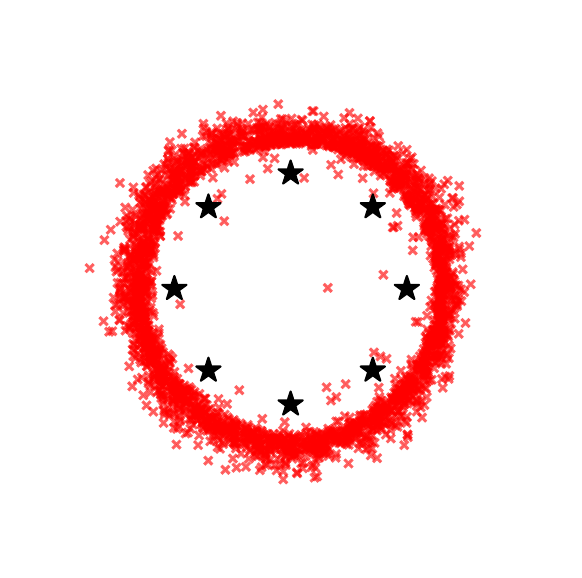}}} & 
            \raisebox{6mm}{\rotatebox{270}{\small Baselines}} \\
            \small SFT & \small RWR & \small DPO & \\
        \end{tabular}
    \end{tabular}

    \caption{\textbf{Qualitative comparison on the 8-Gaussians toy task.} The left panel shows the initial pre-trained Gaussian distribution. The top row illustrates the trajectory of our reward-guided fine-tuning method across 5k, 15k, and 25k iterations. In contrast, the bottom row presents the converged results of baselines at 20k steps.}
    \label{fig:2d_8gaussian}
    \vspace{-14pt}
\end{figure}

\paragraph{\boldmath$W_2$-regularized Distributional Updates.}
Furthermore, the JKO scheme includes the 2-Wasserstein cost as a proximal term, which explicitly penalizes large distribution shifts during fine-tuning. Combined with the steepest descent property of WGF, this induces smooth and controlled distributional updates: the model evolves through small, conservative modifications around the pre-trained distribution while progressively moving toward the reward-weighted target (Figure~\ref{fig:2d_8gaussian}). This minimizing-movement structure has been shown to enable stable and efficient optimization in generative modeling and sampling tasks \citep{asbs, asbm}. Therefore, fine-tuning via the JKO scheme mitigates reward hacking and mode collapse.

\paragraph{Theoretical Justification.} 
The following remark provides a theoretical justification for our fine-tuning framework: under a regularity condition on $\nu_r$, the WGF in Eq.~\eqref{eq:fine-tuning_pde} converges exponentially to the reward-weighted target distribution $\nu_r$. Although $\nu_r$ is generally not log-concave in practical fine-tuning tasks, this guarantee provides theoretical motivation for our training algorithm as follows:
\begin{remark}[Convergence guarantee {\citep[Thm.~11.2.1]{ambrosio}}] \label{rmk:convergence} Suppose $\nu_r$ is $\lambda$-strongly log-concave. Then, choosing the KL divergence as the $f$-divergence, the flow $\rho_t$ converges exponentially in the $\mathcal{W}_2$-distance: \begin{equation} \label{eq:exp_convergence}        
\mathcal{W}_2(\rho_t, \nu_r) \leq C e^{-\lambda t} 
\end{equation} for some constant $C>0$ depending on the initial distribution. 
\end{remark}

\section{Related Works}
\paragraph{One-step Generative Models.} 
One-step generative models aim to reduce the number of inference steps in diffusion/flow-based models, initially by distilling pre-trained multi-step models into few-step variants \cite{salimans2022progressive, meng2023distillation}.
Consistency Models (CM) \cite{song2023consistency} learn a mapping from any point along the ODE trajectory to its endpoint. CM makes it possible to train one-step models from scratch. The Consistency Trajectory Model (CTM) \cite{kim2023consistency} generalizes CM by training a neural network to enforce consistency across different time steps. 
MeanFlow \cite{geng2025mean} learns an average velocity field across time intervals, significantly outperforming other one- or few-step diffusion and flow models.

\paragraph{Reward Fine-Tuning.}
Recent works have explored fine-tuning diffusion and flow models to align with a reward function while preserving the pre-trained distribution. An early approach, DDPO \cite{black2023training}, directly optimizes model parameters for arbitrary rewards but often suffers from reward hacking, i.e., overfitting to a few high-reward samples. To mitigate this, DPOK \cite{fan2023dpok} incorporates KL regularization to preserve diversity. Similarly, ORW-CFM \cite{fan2025online} incorporates the Wasserstein distance to prevent policy collapse and maintain diversity in the fine-tuning of flow models. However, since these methods rely on iterative diffusion or flow-based generation processes, they cannot be directly applied to fine-tuning of one-step models. 

\section{Experiments}
In this section, we evaluate the effectiveness of our method across diverse settings and reward functions.

\begin{table}[t]
    \centering
    \setlength{\tabcolsep}{3.5pt}
    \small
    \begin{tabular}{l cc cc cc}
    \toprule
    \multirow{2}{*}{\textbf{Method}}
    & \multicolumn{2}{c}{\textbf{Incomp.}}
    & \multicolumn{2}{c}{\textbf{Comp.}}
    & \multicolumn{2}{c}{\textbf{Class 5}}\\
    \cmidrule(lr){2-3}\cmidrule(lr){4-5}\cmidrule(lr){6-7}
    & Rwd $\uparrow$ & FID $\downarrow$ & Rwd $\uparrow$ & FID $\downarrow$
    & Rwd $\uparrow$ & FID $\downarrow$\\
    \midrule
    \textit{Pre-trained} & 1.30 & 2.90 & -1.30 & 2.90 & -7.71 & 2.90 \\
    \midrule
    SFT  & 0.96 & 109.8 & -0.89 & 163.4 & -3.82 & 31.3\\
    RWR  & 0.92 & 162.1 & \textbf{-0.86} & 175.1 & -5.56 & 35.3\\
    DPO  & 1.36 & 17.50  & -1.28 & 60.30  & -8.18 & 2.90\\
    \textbf{Ours} ($\beta=20$) & \textbf{1.62} & 55.9 & -0.95 & 161.9
    & \textbf{-0.87} & 67.70\\
    \bottomrule
    \end{tabular}
    \caption{
    \textbf{Quantitative Comparison across Reward Tasks on CIFAR-10.}
    We report the final Reward (Rwd) and FID scores for non-differentiable (Incompressibility/Incomp. and Compressibility/Comp.) and differentiable (Class 5) tasks. FID is computed with 50k samples.
    }
    \label{tab:cifar_unified}
    \vspace{-8pt}
\end{table}

\begin{figure}[t]
    \centering
    \small
    \setlength{\tabcolsep}{1pt}
    \renewcommand{\arraystretch}{0.5}
    \begin{tabular}{c c c}
        & \textbf{Incompressibility} & \textbf{Class 5: Dog} \\
        \raisebox{8pt}{\smash{\rotatebox{90}{\scriptsize Pre}}} &
        \includegraphics[width=0.43\linewidth]{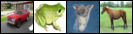} &
        \includegraphics[width=0.43\linewidth]{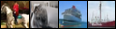} \\
        \raisebox{8pt}{\smash{\rotatebox{90}{\scriptsize SFT}}} &
        \includegraphics[width=0.43\linewidth]{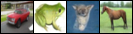} &
        \includegraphics[width=0.43\linewidth]{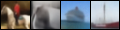} \\
        \raisebox{8pt}{\smash{\rotatebox{90}{\scriptsize RWR}}} &
        \includegraphics[width=0.43\linewidth]{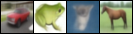} &
        \includegraphics[width=0.43\linewidth]{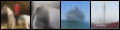} \\
        \raisebox{8pt}{\smash{\rotatebox{90}{\scriptsize DPO}}} &
        \includegraphics[width=0.43\linewidth]{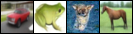} &
        \includegraphics[width=0.43\linewidth]{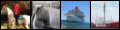} \\
        \raisebox{8pt}{\smash{\rotatebox{90}{\scriptsize \textbf{Ours}}}} &
        \includegraphics[width=0.43\linewidth]{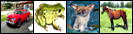} &
        \includegraphics[width=0.43\linewidth]{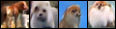} \\
    \end{tabular}
    \caption{\textbf{Qualitative results of generated samples on CIFAR-10.} Left: Incompressibility. Right: Class 5 ("Dog").}
    \label{fig:cifar_samples}
    \vspace{-8pt}
\end{figure}

\subsection{Experimental Setup}

We evaluate our method on a 2D synthetic dataset, CIFAR-10 \cite{Krizhevsky2009LearningML}, and ImageNet 256$\times$256 \cite{deng2009imagenet} to cover both low- and high-dimensional settings. We use a pre-trained MeanFlow model \citep{geng2025mean} as our one-step generator $T_{0}$ with KL divergence. 
We use a diverse set of reward functions to measure fine-tuning performance. To evaluate fidelity, we employ the Fréchet Inception Distance (FID) \cite{heusel2017gans} on both CIFAR-10 and ImageNet. For ImageNet, we further adopt the Aesthetic score \cite{schuhmann2022laion} as our main fidelity measure.

\begin{table}[!t]
    \centering
    \small
    \setlength{\tabcolsep}{3pt}
    \begin{tabular}{clccc}
        \toprule
        Model & \multicolumn{2}{c}{Algorithm} & Reward (KB) $\uparrow$ & FID $\downarrow$ \\
        \midrule
        \multirow{7}{*}{DDPM}
        & \multicolumn{2}{c}{\textit{Pre-trained}} & 1.10 & 29.90 \\
        \cmidrule(lr){2-5}
        & \multicolumn{2}{c}{q-Learning} & \multirow{2}{*}{1.20} & 26.64 \\
        & \multicolumn{2}{c}{DPOK}       &                       & 77.94 \\
        \cmidrule(lr){2-5}
        & \multicolumn{2}{c}{q-Learning} & \multirow{2}{*}{1.30} & 50.58 \\
        & \multicolumn{2}{c}{DPOK}       &                       & 107.99 \\
        \cmidrule(lr){2-5}
        & \multicolumn{2}{c}{q-Learning} & \multirow{2}{*}{1.40} & 89.79 \\
        & \multicolumn{2}{c}{DPOK}       &                       & 138.89 \\        \midrule
        \multirow{5}{*}{MeanFlow}
        & \multicolumn{2}{c}{\textit{Pre-trained}} & 1.29 & 3.90 \\
        \cmidrule(lr){2-5}
        & \multirow{4}{*}{\textbf{Ours}} & $\beta = 1$ & 1.33 & 5.43 \\
        &                       & $\beta = 3$ & 1.42 & 9.32 \\
        &                       & $\beta = 5$ & 1.53 & 24.99 \\
        &                       & $\beta = 20$ & 1.62 & 56.96 \\
        \bottomrule
    \end{tabular}
    \caption{\textbf{Comparison with multi-step fine-tuning on CIFAR-10 (JPEG incompressibility).}
    All DDPM rows are reported from \citet{gao2025reward}, which fine-tunes a DDPM restricted to 20 sampling steps. In this task, FID is computed with 20k samples following the evaluation protocol of \citet{gao2025reward}. Note that the two pre-trained models differ in FID (3.90 vs. 29.90).
    }
    \label{tab:merged_comparison}
    \vspace{-15pt}
\end{table}

\begin{table}[t]
    \centering
    \small
    \setlength{\tabcolsep}{9pt}
    \begin{tabular}{@{}l@{\hskip 16pt}l ccc@{}}
        \toprule
        Task & Method & Rwd. $\uparrow$ & Aes. $\uparrow$ & FID $\downarrow$ \\
        \midrule
        \multirow{3}{*}{Incomp.}
            & Pre-trained & 30.98 & 4.67 & 3.41 \\
            & DPO   & 37.87 & 4.25 & 32.89 \\
            & Ours        & \textbf{46.68} & \textbf{4.92} & \textbf{27.57} \\
        \midrule
        \multirow{3}{*}{Comp.}
            & Pre-trained & -30.98 & 4.67 & 3.41 \\
            & DPO   & \textbf{-18.01} & 4.32 & 45.18 \\
            & Ours        & -19.20 & \textbf{4.55} & \textbf{38.92} \\
        \midrule
        \multirow{3}{*}{B\&W}
            & Pre-trained & -6.16 & 4.67 & 3.41 \\
            & DPO   & -8.78 & 4.54 & \textbf{35.36} \\
            & Ours        & \textbf{-4.80} & \textbf{4.64} & 58.61 \\ 
        \midrule
        \multirow{3}{*}{CLIP-red}
            & Pre-trained & 12.83 & 4.67 & 3.41 \\
            & DPO   & 16.01 & 4.34 & \textbf{30.51} \\
            & Ours        & \textbf{18.02} & \textbf{4.61} & 59.68 \\
        \bottomrule
    \end{tabular}
    \caption{\textbf{Comparison of the baseline and our model on ImageNet 256$\times$256.}
    We report the reward (Rwd.), Aesthetic score (Aes.), and FID across four fine-tuning tasks, all computed with 16k generated samples. CLIP-red and B\&W values are scaled by $\times 100$.}
    \label{tab:imgnet_full_comparison}
    \vspace{-8pt}
\end{table}

\paragraph{Rewards.}
We use two non-differentiable and three differentiable rewards, which can be applied to our method without modification.
For non-differentiable rewards, we employ \textit{JPEG incompressibility} and \textit{compressibility} \citep{black2023training} on both CIFAR-10 and ImageNet 256$\times$256. These rewards are computed through a discrete black-box compression algorithm and thus admit no usable gradient. Incompressibility is maximized by high-frequency texture and compressibility by low-detail.

Among the differentiable rewards, the \textit{class probability} on CIFAR-10 shifts the model distribution toward a single class distribution. On ImageNet, we use \textit{CLIP alignment to `red'}
(CLIP-red) and \textit{Black-and-White} (B\&W)~\citep{jacq2026diffusion}. 
CLIP-red is computed as the CLIP \citep{radford2021learning} alignment score between the generated image and the text prompt `red',
and B\&W is the pixel-wise distance between an image and its black-and-white version, averaged over the color channels.
Note that we do not use reward gradients even for these differentiable rewards, so our method directly applies to all of them without modification.

\paragraph{Baselines.}
We adapt the following methods to operate within a single MeanFlow step via a single noise injection \cite{FlowGRPO} for a fair comparison:
\textit{Supervised Fine-Tuning (SFT)}, \textit{Reward-Weighted Regression (RWR)} \cite{peters2007reinforcement}, and \textit{Direct Preference Optimization (DPO)} \cite{rafailov2023direct}. 
These methods do not require reward gradients, allowing them to be evaluated under the same non-differentiable reward settings as our method. 
Briefly, SFT fine-tunes on the highest-reward samples in each batch, RWR reweights the regression loss by exponentiated rewards, and DPO contrastively prefers higher rewards.

\subsection{Reward-guided Fine-tuning} \label{sec:2d_synthetic}

\paragraph{2D Synthetic Data.}
To investigate the training dynamics of reward-guided fine-tuning for one-step generators, we conduct a qualitative evaluation using 2D synthetic data. Specifically, a model pre-trained on a standard Gaussian distribution is fine-tuned toward an 8-Gaussian mixture defined by a multi-modal reward.

As illustrated in Figure~\ref{fig:2d_8gaussian}, our proposed method exhibits stable and gradual training dynamics. Throughout training (from 5k to 25k iterations), our approach robustly avoids structural collapse and progressively separates the distribution to capture all eight distinct target modes with high precision. In contrast, the baseline methods (SFT, RWR, DPO) are formulated to directly converge to the reward-weighted target distribution. In attempting this direct mapping, they struggle to navigate the complex multi-modal reward landscape.

As shown in the bottom row of Figure~\ref{fig:2d_8gaussian}, they typically suffer from severe mode averaging, i.e., converging to a continuous ring connecting the targets, or fail entirely to capture the distinct modes. This highlights that our WGF formulation provides smooth and controlled distributional updates (Section~\ref{sec:adv_wgf}), enabling stable guidance of the one-step generation process under complex reward signals.

\paragraph{CIFAR-10.}

We report the results on CIFAR-10 for both non-differentiable (Incompressibility and Compressibility) and differentiable (Class 5) reward tasks. Note that fine-tuning methods directly requiring reward gradients \cite{clark2024directly} cannot be applied to non-differentiable rewards.

In the non-differentiable tasks, the baselines fail in two distinct ways (Table~\ref{tab:cifar_unified}).
SFT and RWR fail to improve the incompressibility reward, even decreasing from the initial value. Figure~\ref{fig:cifar_samples} shows that our method effectively improves the reward while preserving the semantic structures of the original images. While DPO achieves quantitative improvements, the resulting images are severely corrupted by noise and blurring artifacts.
Table~\ref{tab:merged_comparison} compares our approach with q-Learning and DPOK fine-tuning of a 20-step DDPM \citep{gao2025reward}, which start from a different pre-trained model (reward 1.10, FID 29.90) than our MeanFlow generator (1.29, 3.90). At a comparable reward level of 1.40, our method attains an FID of 9.32 against their 89.79 and 138,89, using a single-step generator.

For the differentiable task (Class 5), we observe that our method moves the pre-trained model toward higher-reward regions than baseline methods. Our method achieves a reward of $-0.87$, a substantial improvement over the pre-trained model's reward of $-7.71$, as shown in Table~\ref{tab:cifar_unified}. In contrast, DPO fails to improve the reward, while SFT and RWR improve the reward at the expense of fidelity. Figure~\ref{fig:cifar_samples} shows that our method is the best at steering random objects toward the target class ("Dog"). The full qualitative results are in the Appendix.

\begin{figure}[t]
    \centering
    \setlength{\tabcolsep}{1pt}
    \renewcommand{\arraystretch}{0}
    \setlength{\lineskip}{0pt}
    \newcommand{\cell}[1]{\raisebox{-0.5\height}[0.5\height][0.5\height]{\includegraphics[width=0.205\columnwidth]{#1}}}
    \begin{tabular}{@{}l cccc@{}}
        \tiny Pre-trained &
        \cell{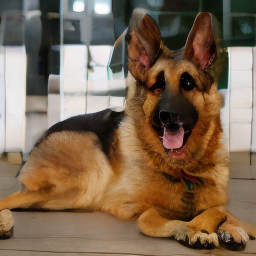} &
        \cell{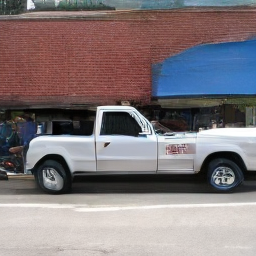} &
        \cell{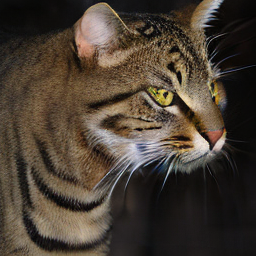} &
        \cell{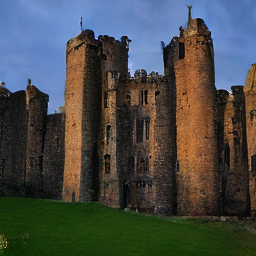} \\
        \noalign{\vspace{2pt}}
        \tiny DPO &
        \cell{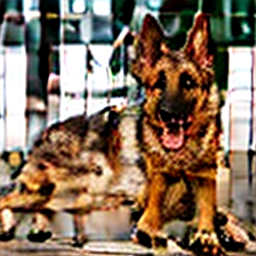} &
        \cell{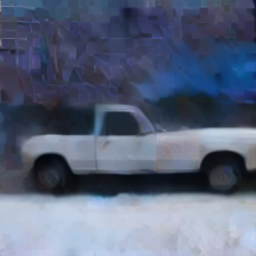} &
        \cell{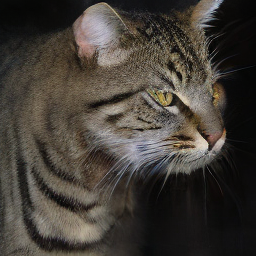} &
        \cell{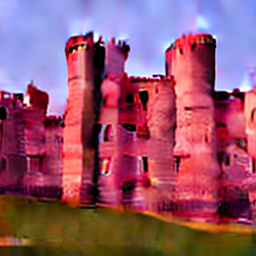} \\
        \noalign{\vspace{2pt}}
        \tiny\textbf{Ours} &
        \cell{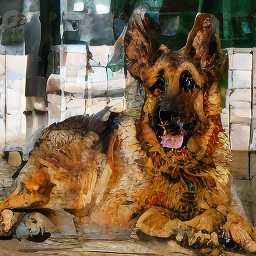} &
        \cell{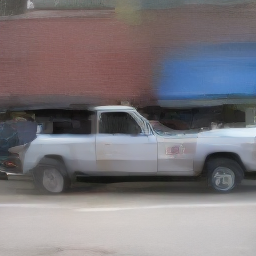} &
        \cell{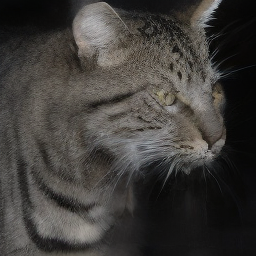} &
        \cell{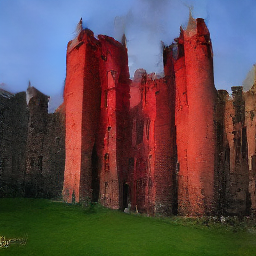} \\
        \noalign{\vspace{2pt}}
        & \tiny (a) Incomp. & \tiny (b) Comp. & \tiny (c) B\&W & \tiny (d) CLIP-red \\
    \end{tabular}
    \caption{\textbf{Qualitative comparisons of generated samples on ImageNet 256$\times$256 across reward targets.}
    Rows correspond to the pre-trained model, the DPO baseline, and our method; columns to different reward targets. All samples use matched class labels and seeds.}
    \label{fig:imgnet_samples}
    \vspace{-14pt}
\end{figure}

\subsection{Scaling to ImageNet 256$\times$256} \label{sec:imagenet}
We further evaluate our framework on ImageNet 256$\times$256 \citep{deng2009imagenet}, using the non-differentiable rewards \textit{JPEG (in)compressibility} and two differentiable rewards \textit{CLIP-red} and \textit{Black-and-White}.
As only DPO improves the incompressibility reward on CIFAR-10 (Table~\ref{tab:cifar_unified}), we adopt DPO as a baseline.
As shown in Table~\ref{tab:imgnet_full_comparison}, where fidelity is measured by aesthetic score and FID, our method attains the highest reward on three of the four tasks and the best aesthetic score on all four tasks.
Therefore, our framework scales directly from the low-dimensional CIFAR-10 setting to the high-dimensional ImageNet 256$\times$256 setting.

\begin{figure}[t!]
  \centering
  \small
  \newcommand{\ylab}{$\vcenter{\hbox{\rotatebox[origin=c]{90}{\scriptsize Reward $\uparrow$}}}$}
  \newcommand{\panel}[1]{$\vcenter{\hbox{\includegraphics[width=0.45\linewidth, height=3.3cm, keepaspectratio=false]{#1}}}$}
  \newcommand{\xlab}{\makebox[0.45\linewidth][c]{\kern 10pt \scriptsize Aesthetic $\uparrow$}}
  \newcommand{\ttl}[1]{\makebox[0.45\linewidth][c]{\kern 10pt \small #1}}
  \begin{tabular}{@{} c@{\hskip 1pt}c @{\hskip 2pt} c@{\hskip 1pt}c @{}}
    & \ttl{(a) Incomp.} & & \ttl{(b) Comp.} \\
    \noalign{\smallskip}
    \ylab & \panel{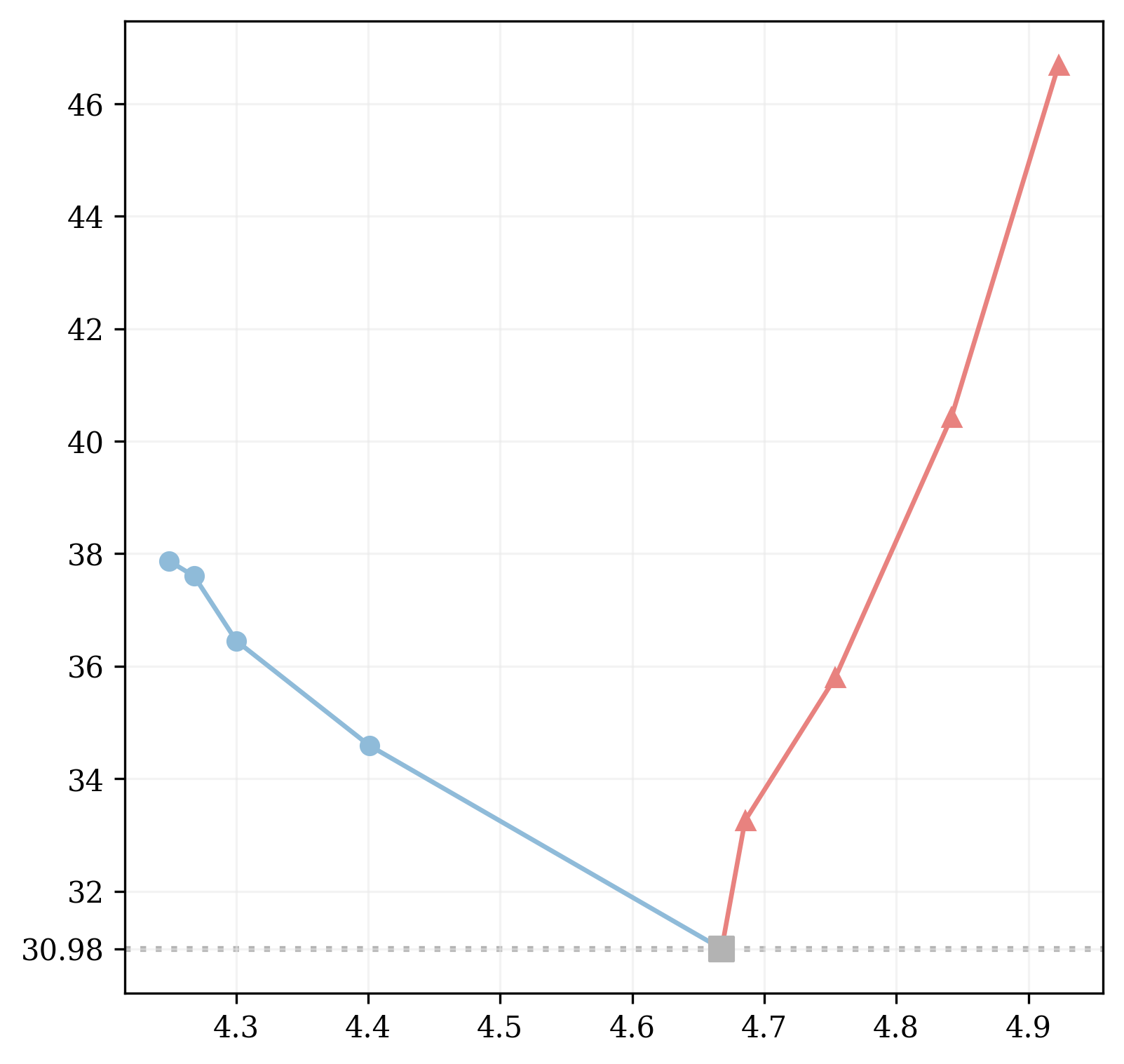} &
    \ylab & \panel{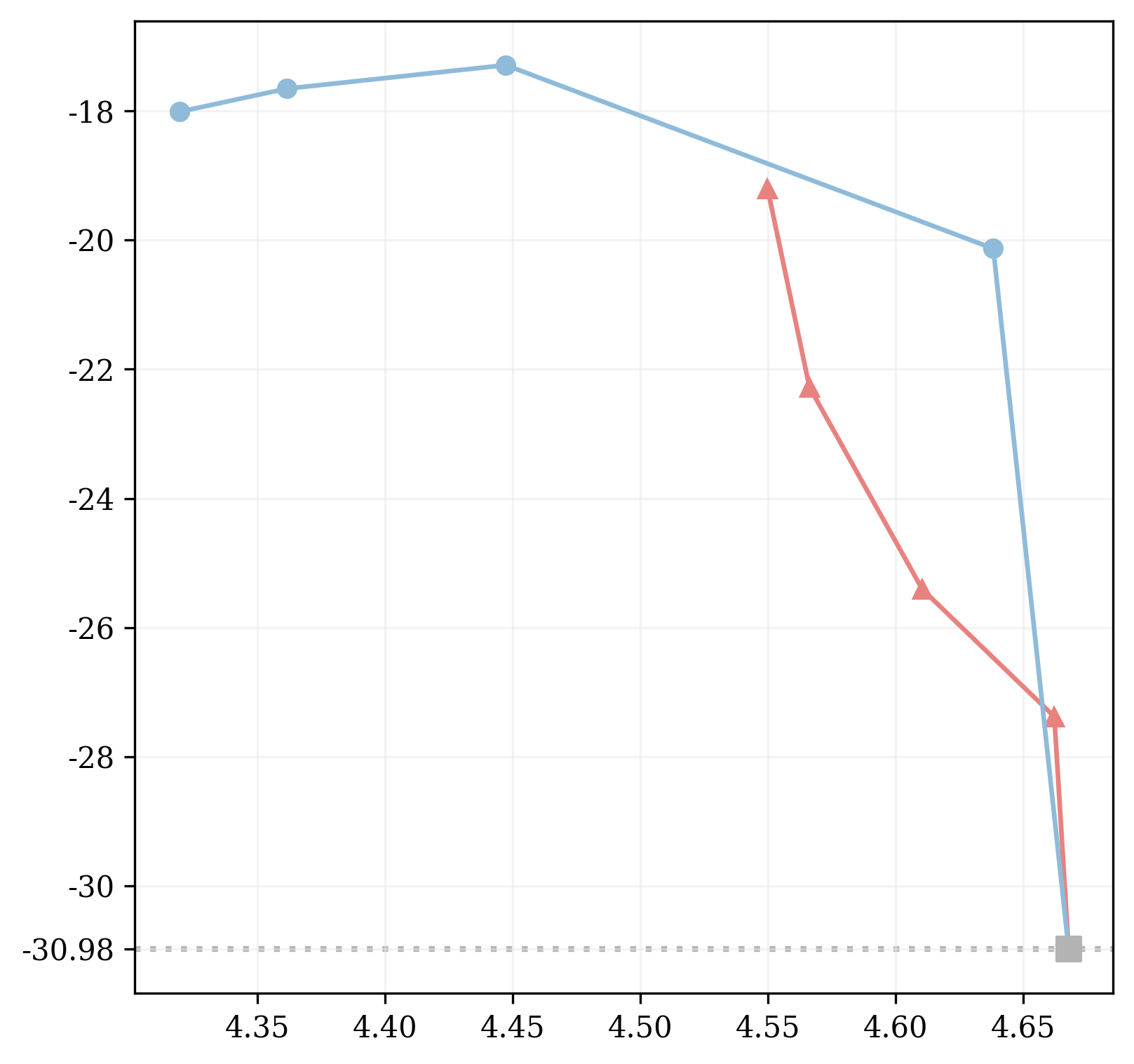} \\
    \noalign{\vspace{1pt}}
    & \xlab & & \xlab \\
    \noalign{\vspace{8pt}}
    & \ttl{(c) B\&W} & & \ttl{(d) CLIP-red} \\
    \noalign{\smallskip}
    \ylab & \panel{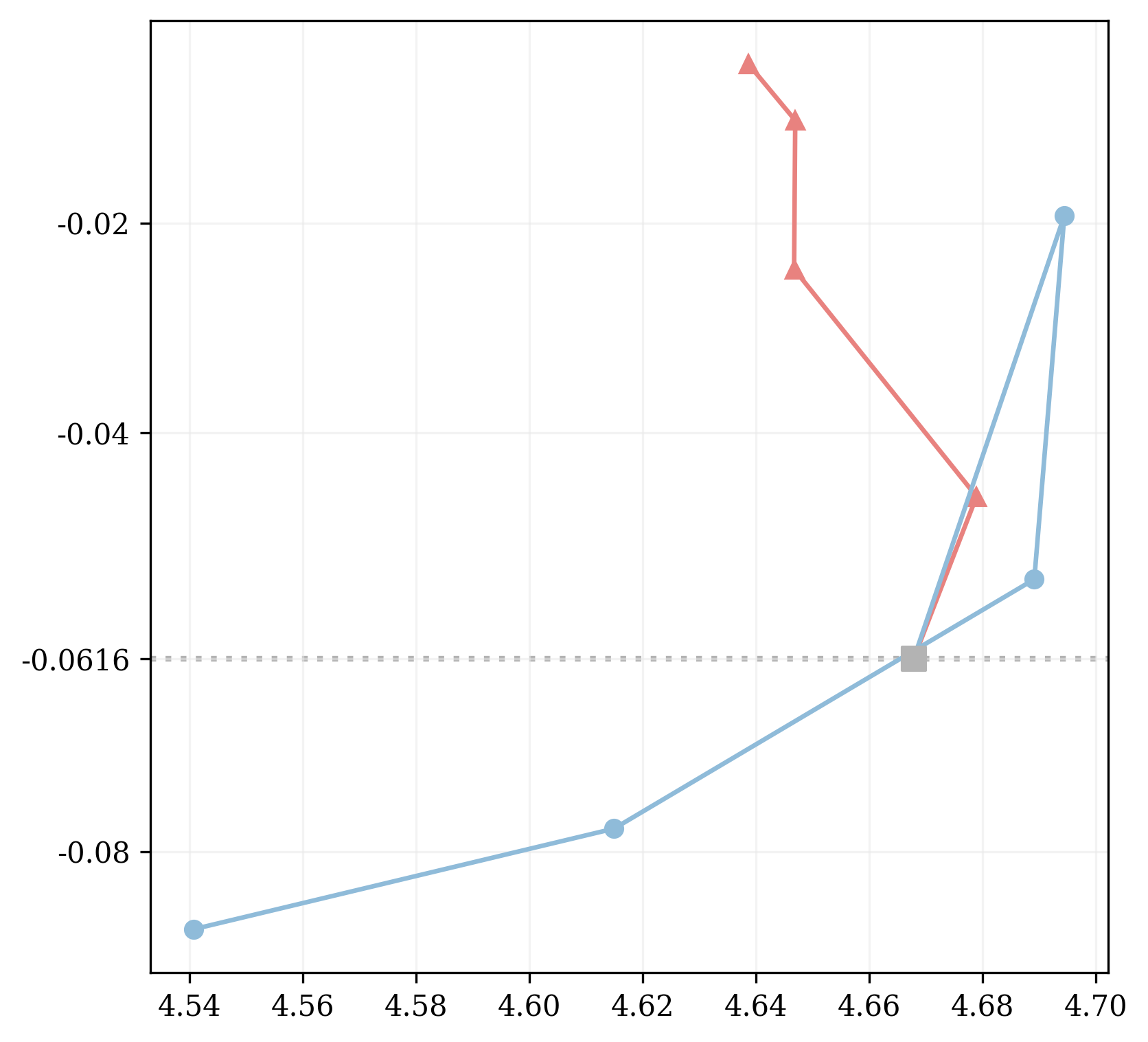} &
    \ylab & \panel{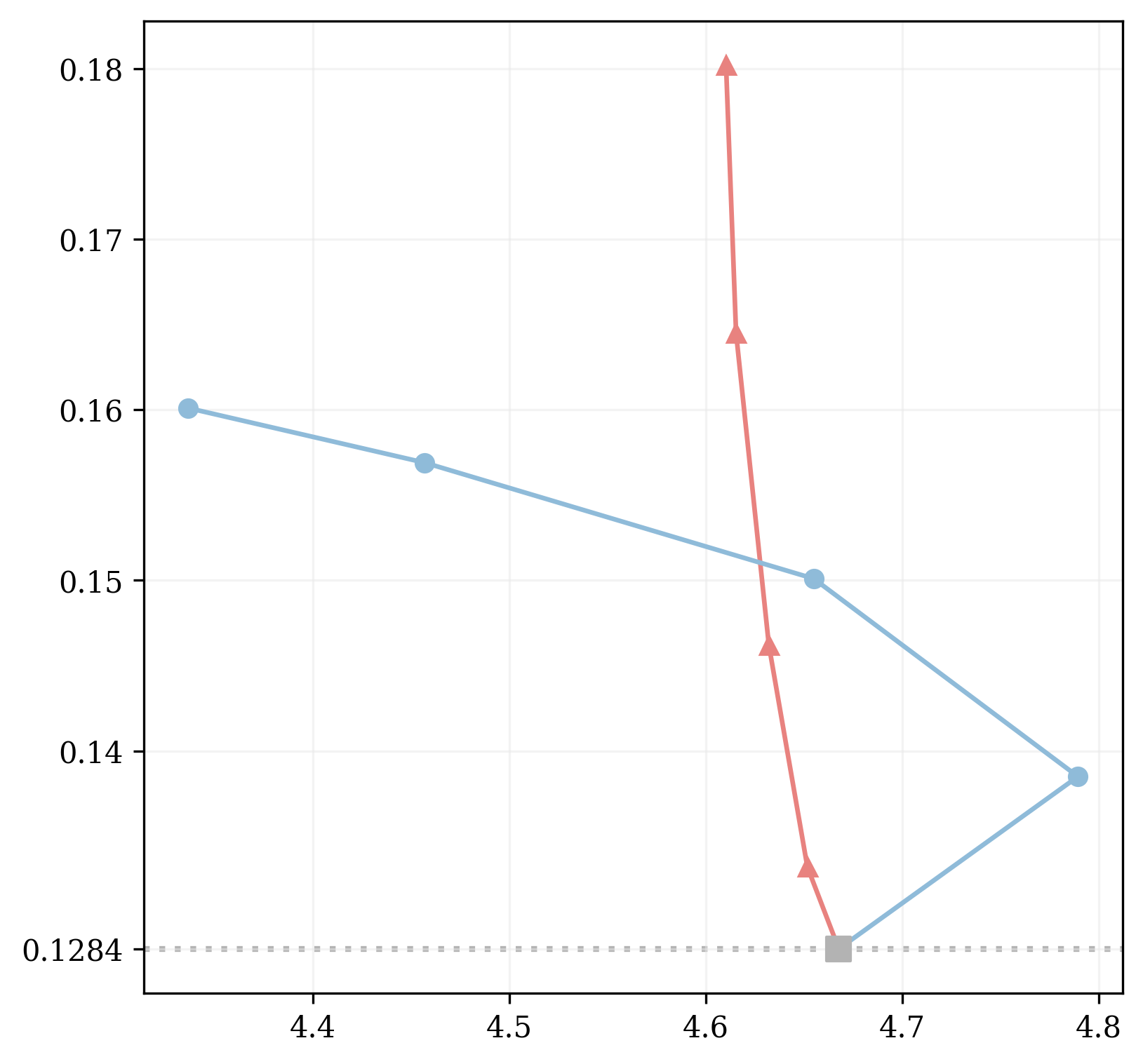} \\
    \noalign{\vspace{1pt}}
    & \xlab & & \xlab \\
  \end{tabular}
  \vspace{4pt}\par
  \includegraphics[width=0.7\linewidth]{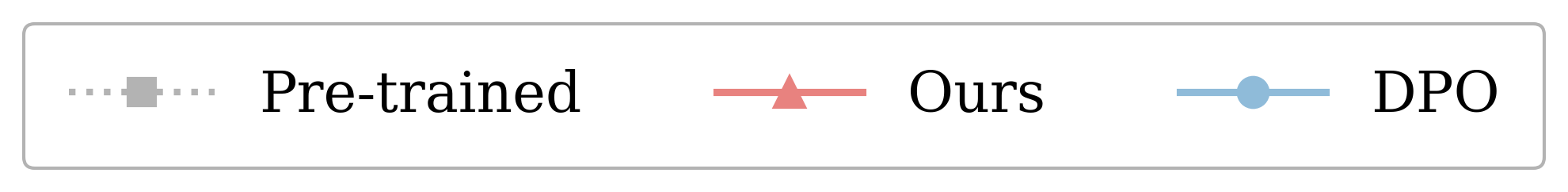}
  \caption{\textbf{Reward--aesthetic trade-off on ImageNet 256$\times$256.} Each panel plots the  reward ($y$-axis) against the Aesthetic score ($x$-axis) over the course of fine-tuning. Markers denote the reward and Aesthetic score at 4k--16k iterations. The upper-right areas indicate a better trade-off.}
  \label{fig:imgnet_pareto_aesthetic}
  \vspace{-15pt}
\end{figure}

For the non-differentiable tasks, our method improves both the reward and the aesthetic score on incompressibility, while DPO raises the reward and loses aesthetic score. On compressibility, DPO reaches a slightly higher reward than ours, but with a large drop in aesthetic score. Figure~\ref{fig:imgnet_samples}(a) and (b) show that DPO adds high-frequency patterns over the whole image on incompressibility and flattens the background on compressibility, while our method changes the texture and preserves the image structure.

For the differentiable tasks, DPO fails in two ways (Figure~\ref{fig:imgnet_pareto_aesthetic}): 
on Black-and-White, its reward peaks at 4k iterations and then falls below the pre-trained value, and on CLIP-red it keeps increasing the reward
while the aesthetic score drops rapidly. 
In contrast, our method maintains the aesthetic score close to the pre-trained value, and also achieves a higher reward on Black-and-White and CLIP-red.

Additionally, our model achieves higher aesthetic scores but also higher FIDs on these two tasks.
The reward-weighted target is far from the data distribution, so improving the reward necessarily increases FID. Our higher FID therefore follows from maximizing the reward better than DPO.
Figure~\ref{fig:imgnet_samples}(c) and (d) show that DPO barely changes the image on Black-and-White and degrades its structure on CLIP-red, while our method edits only the target attribute.
Additional training trajectories and qualitative results for all rewards are provided in the Appendix~\ref{app:additional_results}.

\subsection{Ablation Study} \label{sec:ablation_study}
We conduct two ablation studies on the CIFAR-10 dataset: we first examine the effect of the reward intensity parameter $\beta$, and then analyze our choice between a fixed and an updated pre-trained distribution $\nu$ during training.

\paragraph{The Effect of $\beta$.} 
The reward intensity parameter $\beta$ controls the weight of the target reward during the optimization process. Table~\ref{tab:merged_comparison} sweeps $\beta \in \{1,3,5,20\}$ on a non-differentiable reward JPEG incompressibility task. The reward increases monotonically from 1.33 to 1.62, all above the pre-trained value of 1.29, while FID gradually increases.

\paragraph{\textit{Fixed} vs.\ \textit{Updated} $\nu$.}
We compare the two choices of pre-trained distribution $\nu$ used to evaluate the reward-weighted term in Eq.~\ref{eq:is}, introduced in Section~\ref{sec:jko_reformulation}: \textit{Fixed}, which samples from the pre-trained distribution $\nu$, and \textit{Updated}, which samples from the updated pre-trained distribution $\nu_k$ during training. In practice, the \textit{Updated} approach allows the reward-weighted target to remain concentrated on high-reward regions under the current generator while the \textit{Fixed} approach remains anchored at the pre-trained distribution $\nu$.

\begin{figure}[t!]
  \centering
  \small
  \newcommand{\ylab}{$\vcenter{\hbox{\rotatebox[origin=c]{90}{\scriptsize FID $\downarrow$}}}$}
  \newcommand{\panel}[1]{$\vcenter{\hbox{\includegraphics[width=0.58\linewidth, height=3.6cm, keepaspectratio=false]{#1}}}$}
  \newcommand{\xlab}{\makebox[0.58\linewidth][c]{\kern 10pt \scriptsize Reward $\uparrow$}}
  \newcommand{\ttl}[1]{\makebox[0.58\linewidth][c]{\kern 10pt \small #1}}
  \begin{tabular}{@{} c@{\hskip 1pt}c @{}}
    & \ttl{\textit{Fixed} vs. \textit{Updated}} \\
    \noalign{\vspace{3pt}}
    \ylab & \panel{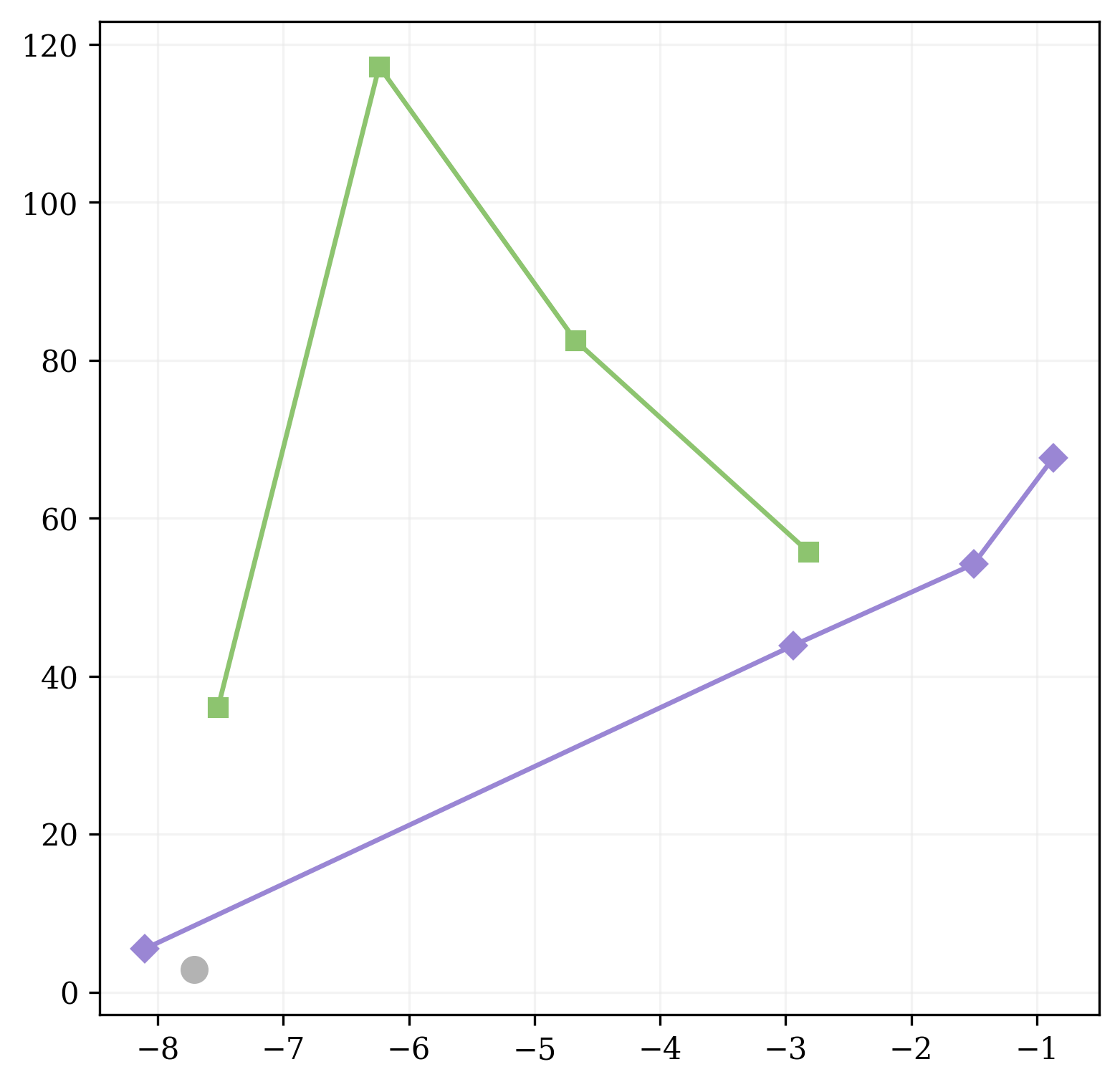} \\
    \noalign{\vspace{1pt}}
    & \xlab \\
  \end{tabular}
  \vspace{4pt}\par
  \includegraphics[width=0.48\linewidth]{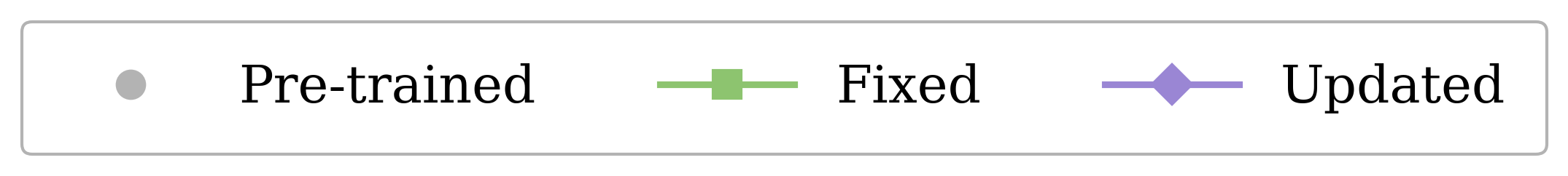}
  \caption{\textbf{Comparison results of \textit{Fixed} vs.\ \textit{Updated} $\nu$.} \\ On CIFAR-10 with the class probability reward for the "Dog" class, the plot shows the reward--fidelity trade-off over the course of fine-tuning under the two choices of pre-trained distribution $\nu$.}
  \label{fig:fixed_vs_updated}
  \vspace{-14pt}
\end{figure}

Figure~\ref{fig:fixed_vs_updated} confirms this using the class reward ("Dog"). \textit{Updated} traces a smooth trajectory in which the reward increases steadily while FID grows only gradually. \textit{Fixed} instead spikes to a much larger FID around 3000 iterations before recovering, and its reward improves far more slowly. Over the whole trajectory, the curve of \textit{Updated} dominates that of \textit{Fixed}, reaching a higher reward at any comparable FID and a lower FID at any comparable reward. Qualitative results are presented in the Appendix~\ref{app:fixed_ablation}.

\section{Conclusion}
In this paper, we propose the first reward-guided fine-tuning framework for one-step generative models via Wasserstein Gradient Flow (WGF). Since existing fine-tuning methods require the intermediate states of a multi-step sampling trajectory, they cannot be directly applied to one-step generators. Our approach formulates fine-tuning as a WGF that evolves the pre-trained distribution toward a reward-weighted target and discretizes this flow through the JKO scheme. The resulting objective enables stable distributional updates that mitigate reward hacking, and it requires no reward gradients, so it applies to non-differentiable rewards. Experiments demonstrate superior reward alignment over baseline approaches on both synthetic and real datasets. As a limitation, we do not evaluate our method on prompt-conditional text-to-image generation, an important application of reward-based fine-tuning. Extending our framework to this setting is a promising direction for future work.

\section*{Acknowledgments}
This work was supported by the National Research Foundation of Korea (NRF) grant funded by the Korea government (MSIT) (RS-2024-00349646, RS-2026-25506577), National Institute for Mathematical Sciences (NIMS) grant funded by the Korea government (MSIT) (No. S26120000), and KIAS Individual Grant (AP102301, AP102303) via the Center for AI and Natural Sciences at Korea Institute for Advanced Study.
This work was also supported by the Center for Advanced Computation at Korea Institute for Advanced Study.

\bibliography{mybib}

\newpage
\clearpage
\appendix
\input{appendix}

\end{document}

%% file: appendix.tex
\section{Algorithm} \label{app:algorithm}

We provide a detailed algorithm in Algorithm \ref{alg:finetune}.

\begin{algorithm}[htbp]
\caption{Reward-Guided Finetuning via WGF}
\label{alg:finetune}
    \begin{algorithmic}[1]
    \REQUIRE Pretrained generator $T_{0}$, discriminator network $v_{\phi}$, reward function $r(\cdot)$, smoothing rate $\gamma$ and intensity parameter $\beta > 0 $
    \STATE Initialize $T_{\text{old}} \leftarrow T_{0}$, $T_\theta \leftarrow T_0$
    \STATE $M \leftarrow \text{mini-batchs} \sum_{y \sim T_{\text{old}}}\exp(\beta \cdot r(y))$
    
    \FOR{$k = 0, 1, 2 , \dots, K$}
        \FOR{$i = 0, 1, 2 , \dots, N$}
            \STATE Sample mini-batches $z \sim \mathcal{N}(0,I)$
            \STATE Let $y = T_{\text{old}}(z)$
            \STATE Compute rewards $\{ r(y) \}$ and update the normalization term:
            \[
            M \leftarrow (1 - \gamma) M + \gamma \cdot \text{mean}\{ \exp(\beta \cdot r(y)) \}
            \]
            \STATE Generate samples $\hat{y} = T_{\theta}(z)$
            \STATE \textbf{Discriminator update:} minimize
            \[
            \mathcal{L}_{v} = v_{\phi}(\hat{y}) - f^{\circ}\!\left(v_{\phi}(y)\right) \frac{\exp(\beta \cdot r(y))}{M}
            \]
            \STATE Sample $z \sim \mathcal{N}(0,I)$ and compute $y = T_{\text{old}}(z)$, $\hat{y} = T_{\theta}(z)$, 
            \STATE \textbf{Generator update:} minimize
            \[
            \mathcal{L}_{T} = c\!\left(y, \hat{y}\right) - v_{\phi}(\hat{y})
            \]
        \ENDFOR
        \STATE Update reference generator: $T_{\text{old}} \leftarrow T_{\theta}$
    \ENDFOR
    \end{algorithmic}
\end{algorithm}

\section{Implementation Details} \label{app:imple_details}

\subsection{Rewards} \label{app:reward_defs}
We evaluate our method on diverse reward functions, including both differentiable and non-differentiable objectives. We use JPEG (in)compressibility~\citep{black2023training} on both CIFAR-10 and ImageNet 256$\times$256. For CIFAR-10, we also test a class probability reward. For ImageNet, we further evaluate on CLIP alignment `red', Black-and-white~\citep{jacq2026diffusion}.

\paragraph{JPEG Incompressibility and Compressibility.}
Following \citet{black2023training}, we use \textit{JPEG (in)compressibility} as non-differentiable reward tasks.  Incompressibility is defined as the file size (KB) of the generated image saved in JPEG format, which encourages synthesizing high-frequency details. Conversely, compressibility is formulated as the negative of the JPEG file size. Since JPEG compression algorithm is a discrete, black-box operation, these rewards are strictly non-differentiable, making standard gradient-based optimization inapplicable. Note that such non-differentiable rewards are common in practice \citep{fan2023dpok}, yet they cannot be handled by methods that backpropagate through the reward \citep{clark2024directly, prabhudesai2023aligning}.

\paragraph{Class Probability.}
We use the \textit{class probability of the `dog' class (class 5)} in CIFAR-10 as a differentiable reward. Given a pretrained ResNet-56 classifier $p_\phi$, the reward is defined as $r(y) = \log p_\phi\big(c = 5 \,\big|\, y\big).$ This evaluates whether the model can conditionally shift towards a specific semantic mode.


\paragraph{CLIP Alignment `red'.}
Following the RMMD~\citep{jacq2026diffusion} framework, we measure \textit{CLIP \citep{radford2021learning} alignment using a single keyword}. With a frozen CLIP ViT-L/14 model, the reward is computed as the cosine similarity between the image embedding and the text embedding of the word ``red''. This directly drives the image color toward the target hue.

\paragraph{Black and White.}
Also adopted from \cite{jacq2026diffusion}, the \textit{Black-and-White reward} encourages grayscale images by penalizing the $L_1$ variance across RGB channels: $r(x) = -\frac{1}{CHW} \sum_{c,i,j} \left\vert{} x_{c,i,j} - \mu_{i,j} \right\vert{}$, where $x \in [0,1]^{C \times H \times W}$ is the normalized generated image, and $\mu_{i,j}$ is the mean across channels at spatial location $(i,j)$. This drives the image toward zero saturation in pixel space.

\subsection{Baselines} \label{app:baseline_defs}
\paragraph{One-Step Generative Setting.}

We compare our method with three representative fine-tuning baselines. 
Throughout, $T_0$ denotes the pre-trained one-step generator, $T_\theta$ the model being fine-tuned, and $T_{\text{old}}$ the reference generator, which is initialized at $T_0$ and updated to the current $T_\theta$ during training, as in Algorithm~\ref{alg:finetune}.
Let $z_i, \epsilon_i \sim \mathcal{N}(0, I)$ denote independent Gaussian samples, where $z_i$ is the input to the generator and $\epsilon_i$ is used for noise injection.

\paragraph{Candidate Generation.}
At each update, we sample $G$ initial noise $\{z_i\}_{i=1}^{G} \sim \mathcal{N}(0, I)$, where $G$ denotes the
group size, and generate a candidate sample $x_i$ from each $z_i$ as
follows. 
For CIFAR-10, we construct an intermediate latent at
$t^\ast \sim \mathcal{U}[t_{\min}, t_{\max}]$ by injecting noise:
\begin{equation}
    z_{\text{inter},i} = (1 - t^\ast)\, T_{\text{old}}(z_i)
    + t^\ast \epsilon_i,
    \quad \epsilon_i \sim \mathcal{N}(0, I),
\end{equation}
and generate the candidate $x_i$ from $z_{\text{inter},i}$ via a
single MeanFlow velocity step from $t=t^\ast$ to $t=0$.
For ImageNet 256$\times$256, we instead perturb each initial noise
with scale $\sigma_{\exp}$,
\begin{equation}
    z_i' = \sqrt{1 - \sigma_{\exp}^2}\, z_i + \sigma_{\exp}\, \epsilon_i,
    \quad \epsilon_i \sim \mathcal{N}(0, I),
\end{equation}
and obtain the candidate as $x_i = T_{\text{old}}(z_i')$.
In both cases, this procedure yields a candidate set
$\{(z_i, x_i)\}_{i=1}^{G}$ with rewards $r(x_i)$.

\paragraph{Baseline Implementation.}
We write $(z^{\pm}, x^{\pm}) := (z_{i^{\pm}}, x_{i^{\pm}})$ for the highest- and lowest reward candidates in the group, where the corresponding indices are given by $i^{+} = \arg\max_{i \in \{1, \dots, G\}} r(x_i)$ and $i^{-} = \arg\min_{i \in \{1, \dots, G\}} r(x_i)$.
SFT uses the highest-reward pair $(z^{+}, x^{+})$, while DPO
contrasts $(z^{+}, x^{+})$ against $(z^{-}, x^{-})$.

The objective for each baseline is defined as follows:

\begin{itemize}

    \item \textbf{Supervised Fine-Tuning (SFT)} employs two distinct target selection strategies depending on the reward objective:
    \begin{itemize}
        \item \textit{Winner Strategy (for incompressibility and compressibility):} Selects the single highest-reward sample $x^{+}$ and optimizes the direct reconstruction objective:
        \begin{equation}
            \mathcal{L}_{\text{SFT-Winner}} = \mathbb{E} \left[ \| T_\theta(z^{+}) - x^+ \|^2 \right].
        \end{equation}
        
        \item \textit{Top-$k$ Mean Strategy (for class 5):} Constructs a reward-weighted blended target from the top-$k$ highest-reward samples $\{x_i\}_{i=1}^k$. It minimizes the distance between the mean predicted output and this blended target:
        \begin{equation}
            \mathcal{L}_{\text{SFT-TopK}} = \mathbb{E} \left[ \Big\| \frac{1}{k} \sum_{i=1}^k T_\theta(z_i) - \sum_{i=1}^k w_i x_i \Big\|^2 \right],
        \end{equation}
        where $w_i = \text{softmax}(\beta \cdot r_i)$ represents the reward-scaled weights for each sample, and $r_i$ is the corresponding reward. We adopt this strategy to mitigate the instability often caused by relying on a single sample.
    \end{itemize}
    
    \item \textbf{Reward-Weighted Regression (RWR)} regresses
    $T_\theta$ onto all $G$ candidates, weighted by their
    exponentiated rewards:
    \begin{equation}
        \begin{aligned}
        \mathcal{L}_{\text{RWR}} &= \mathbb{E} \left[ \sum_{i=1}^G w_i \|T_\theta(z_i)-x_i\|^2\right], \\
        \quad w_i &= \frac{\exp(r(x_i)/\tau)}{\sum_j \exp(r(x_j)/\tau)},
        \end{aligned}
    \end{equation}
    where $\tau$ is a temperature scaling parameter.

    \item \textbf{Direct Preference Optimization (DPO)} directly optimizes the policy by contrasting a higher-reward sample $x^+$ and a lower-reward sample $x^-$. Let
    $\ell_\theta(i) = \| T_\theta(z_i) - x_i \|^2$ and
    $\ell_{\text{old}}(i) = \| T_{\text{old}}(z_i) - x_i \|^2$
    denote the reconstruction losses under the current and reference
    models, respectively. Then
    
    \begin{equation}
        \begin{aligned}
            \mathcal{L}_{\text{DPO}} = -\mathbb{E}[ \log \sigma ( \beta_\text{DPO}  [ (\ell_\theta(i^-) -\ell_{\text{old}}(i^-) ) \\ - ( \ell_\theta(i^+) - \ell_{\text{old}}(i^+)  ) ])],
        \end{aligned}
    \end{equation}
    where $\sigma$ is the sigmoid function and $\beta_\text{DPO}$ controls the strength of the KL penalty.
\end{itemize}

Due to the expensive computation of ImageNet 256$\times$256, we compare with DPO because DPO improves the incompressibility reward on CIFAR-10. 
Since the generator is class-conditional, we draw a class label $c_i$ together with each $z_i$ and write $T(z_i,c_i)$.
We further generate the candidates and compute the anchor term from the fixed pre-trained model $T_0$ rather than from $T_\text{old}$.
The anchor term $\lambda_\text{anc} \|T_\theta(z)-T_0(z)\|^2$, evaluated on a separately drawn batch, keeps $T_\theta$ close to the pre-trained model without collapse.

\begin{figure}[t!]
  \centering
  \small
  \newcommand{\ylab}{$\vcenter{\hbox{\rotatebox[origin=c]{90}{\scriptsize Reward $\uparrow$}}}$}
  \newcommand{\panel}[1]{$\vcenter{\hbox{\includegraphics[width=0.45\linewidth, height=3.3cm, keepaspectratio=false]{#1}}}$}
  \newcommand{\xlab}{\makebox[0.45\linewidth][c]{\kern 10pt \scriptsize Iteration}}
  \newcommand{\ttl}[1]{\makebox[0.45\linewidth][c]{\kern 10pt \small #1}}
  \begin{tabular}{@{} c@{\hskip 1pt}c @{\hskip 2pt} c@{\hskip 1pt}c @{}}
    & \ttl{(a) Incomp.} & & \ttl{(b) Comp.} \\
    \noalign{\smallskip}
    \ylab & \panel{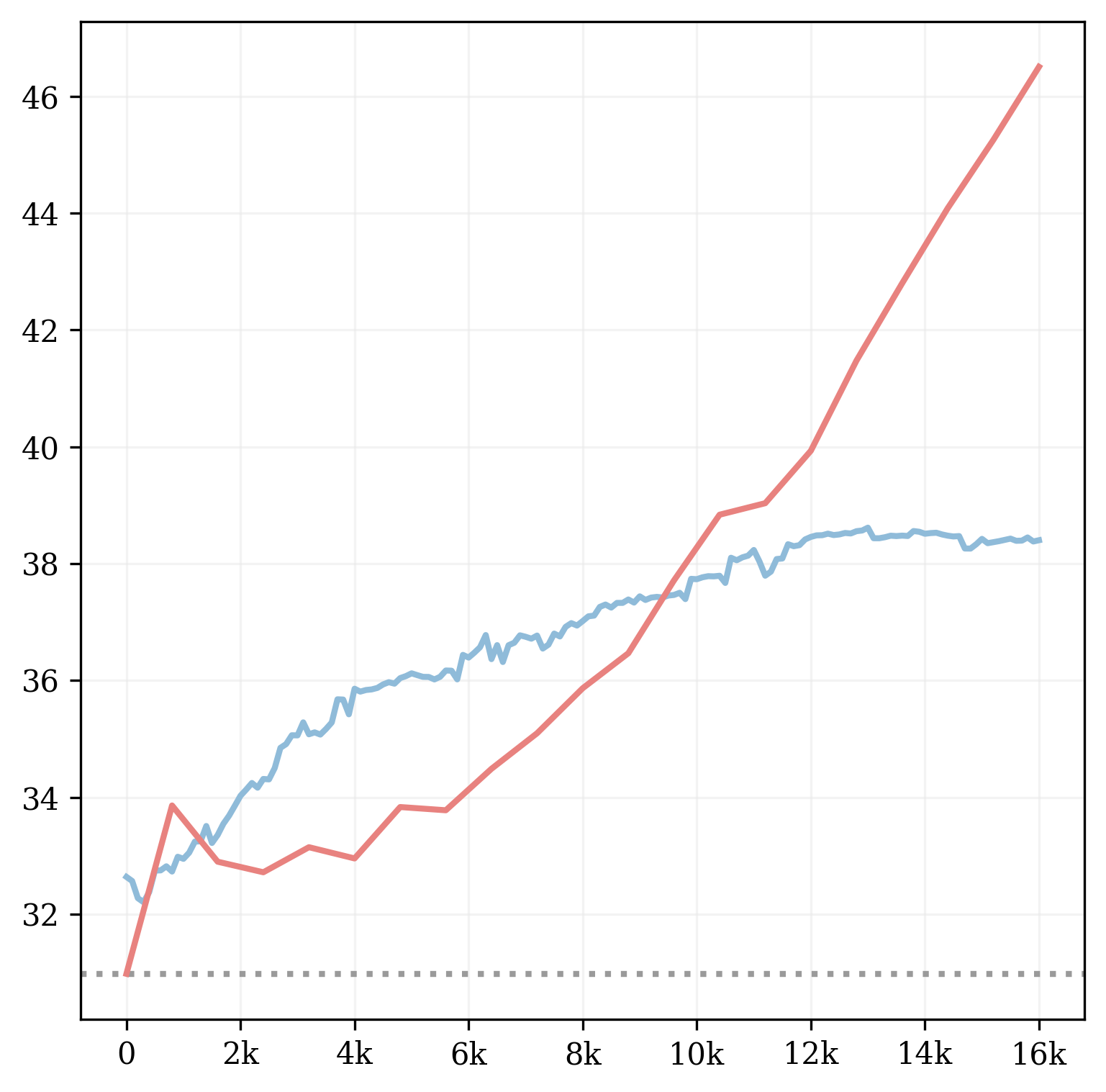} &
    \ylab & \panel{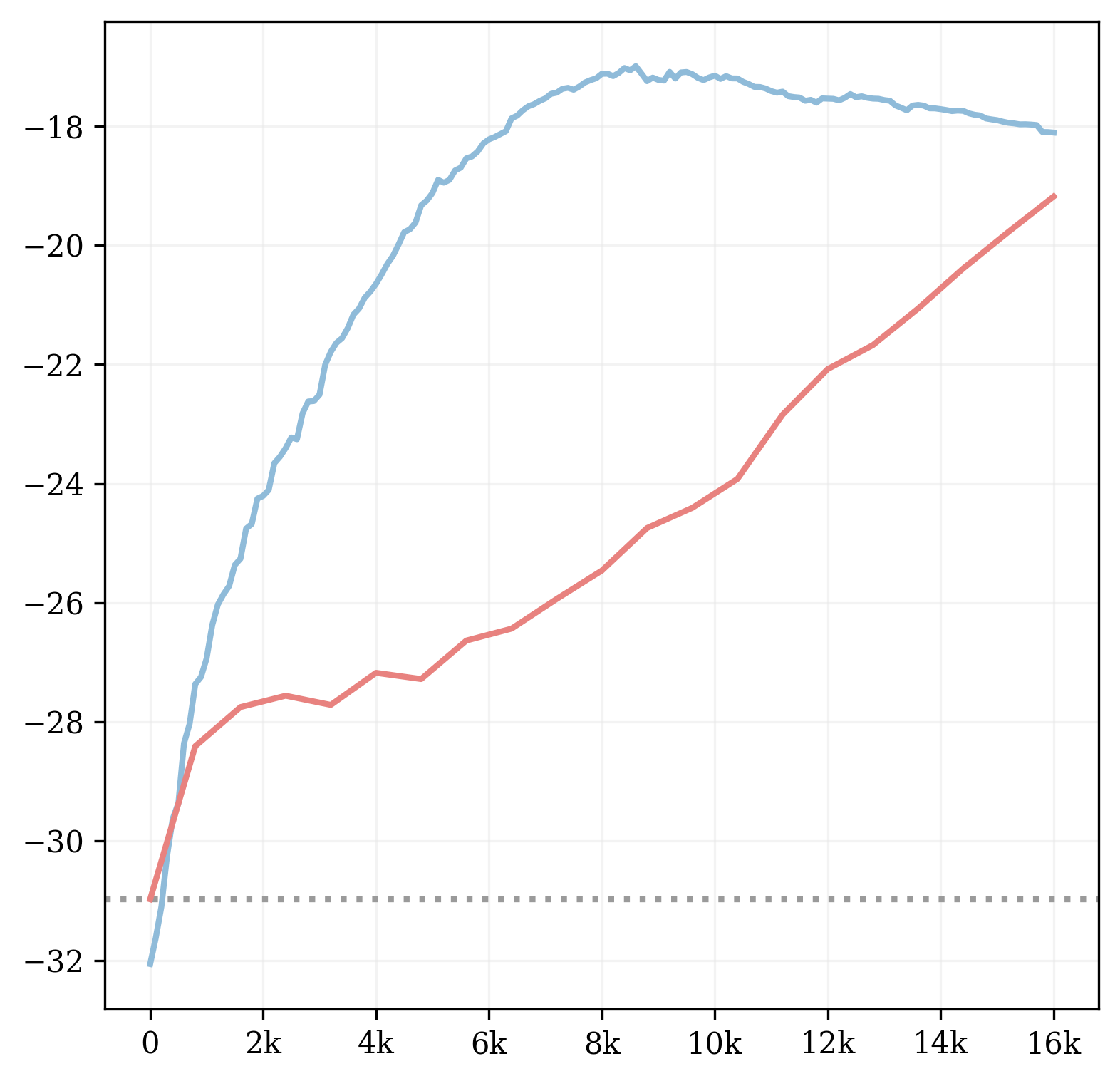} \\
    \noalign{\vspace{1pt}}
    & \xlab & & \xlab \\
    \noalign{\vspace{8pt}}
    & \ttl{(c) B\&W} & & \ttl{(d) CLIP-red} \\
    \noalign{\smallskip}
    \ylab & \panel{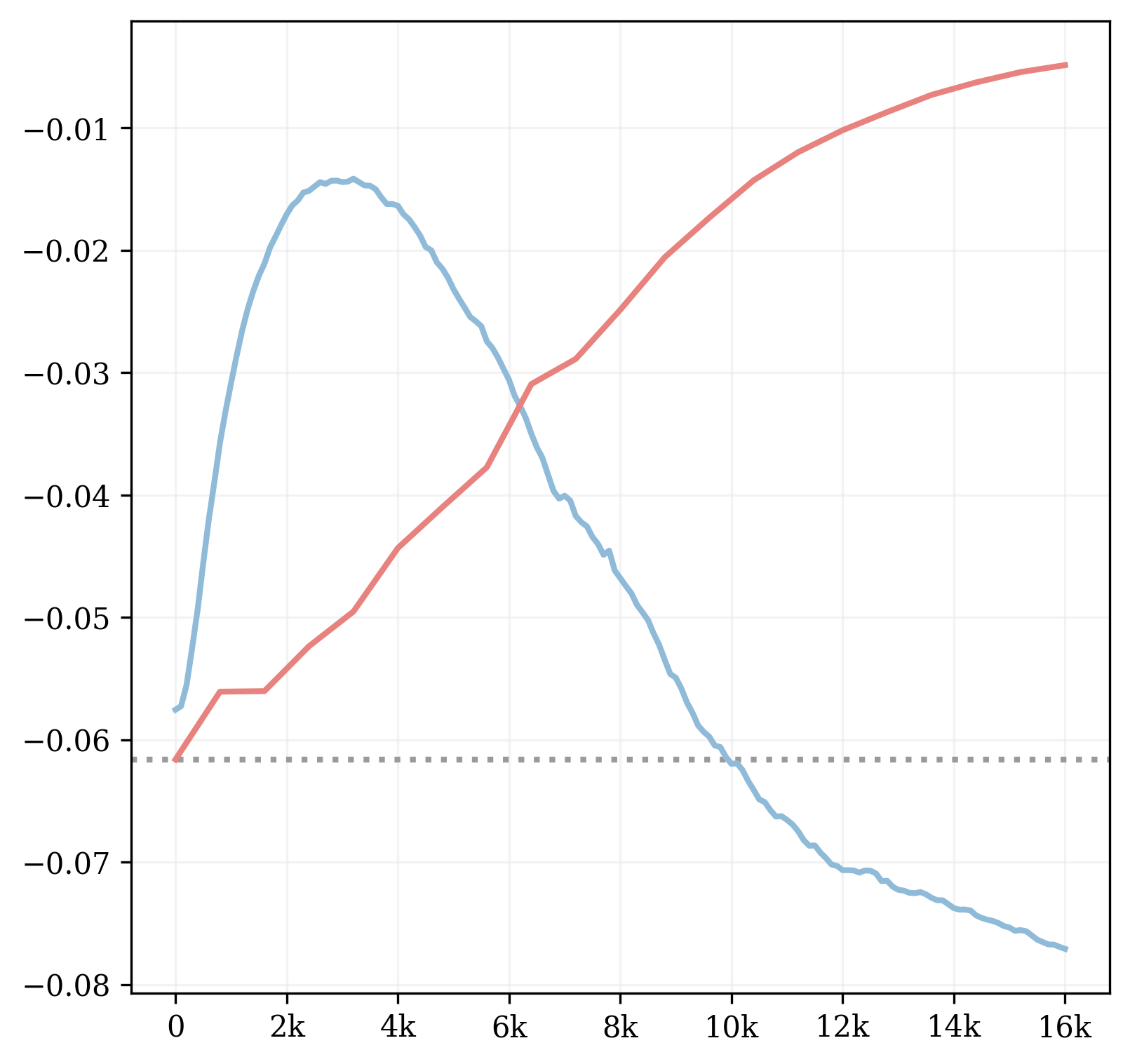} &
    \ylab & \panel{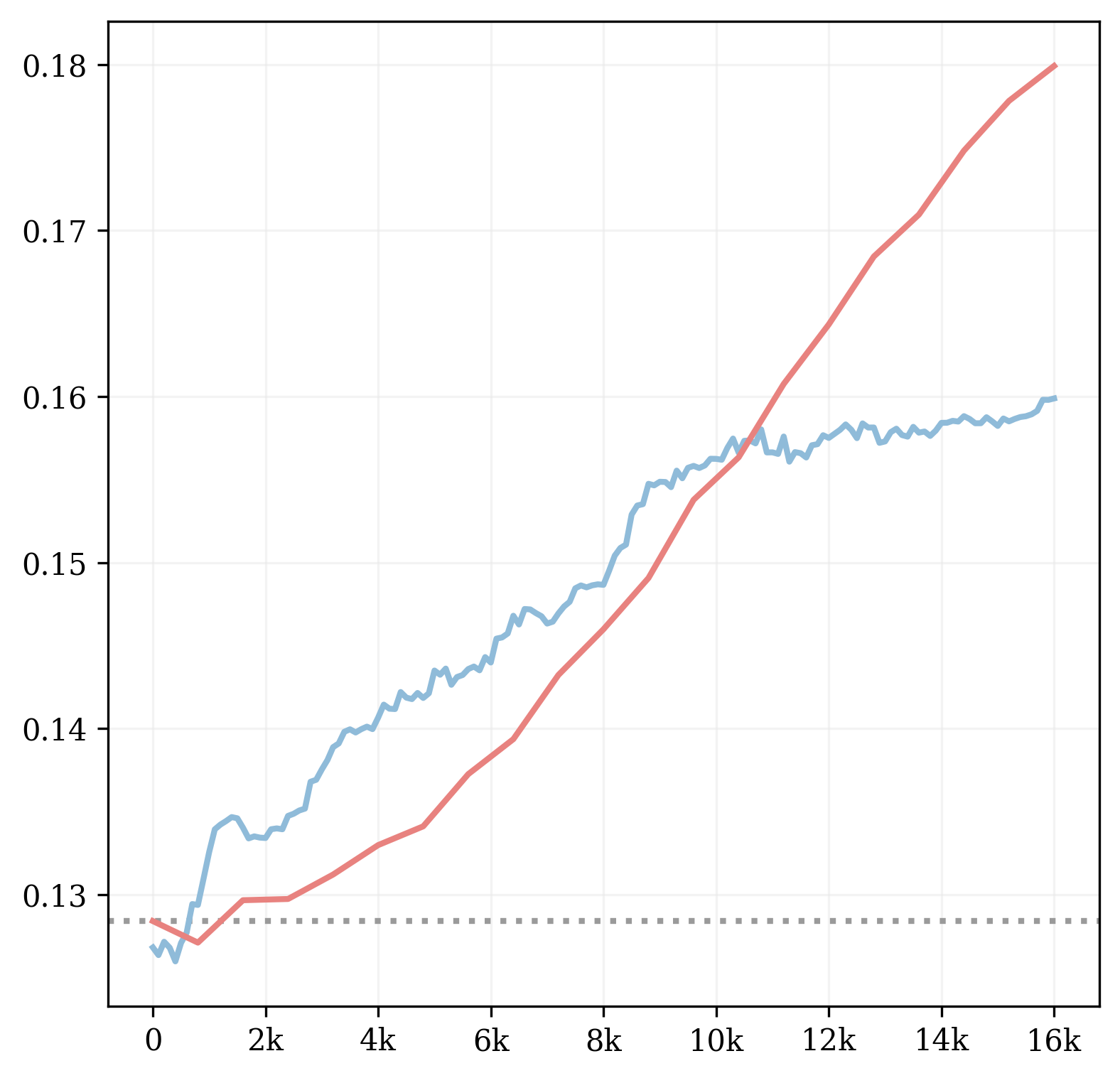} \\
    \noalign{\vspace{1pt}}
    & \xlab & & \xlab \\
  \end{tabular}
  \vspace{4pt}\par
  \includegraphics[width=0.7\linewidth]{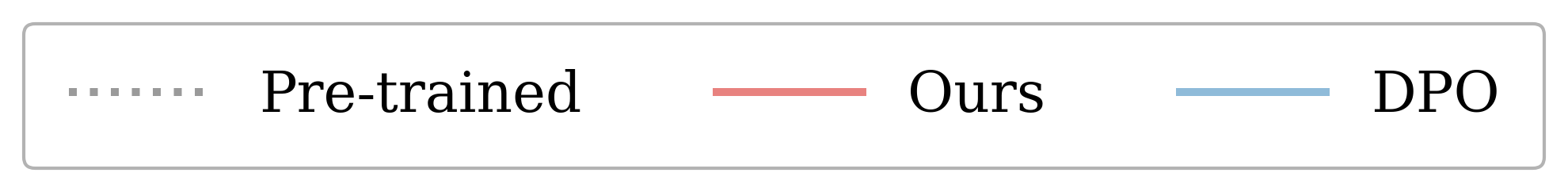}
  \caption{\textbf{Reward curves over training on ImageNet 256$\times$256.} Each panel plots the target reward against the number of fine-tuning iterations for one reward objective.}
  \label{fig:imgnet_reward_curves}
\end{figure}-

\subsection{Training Details}
For all experiments, we evaluate models using the CIFAR-10 32$\times$32 dataset and the high-resolution ImageNet 256$\times$256 dataset. All models are trained using a single NVIDIA A100 GPU. Training uses a fixed seed of 1024; all evaluation sampling uses seed 42, re-seeded before each checkpoint.

\paragraph{CIFAR-10 Experiments.}
For CIFAR-10 32$\times$32, all baseline models and our proposed framework are trained for a total of 5,000 iterations. Specifically, for Class 5 reward experiment, a discriminator warmup phase of 2,000 iterations is applied to ensure stable initial gradient signals before joint optimization. All quantitative results are reported at the final iteration. 
While FID is generally computed using 50k samples at 5,000 iterations, we follow the evaluation protocol of \citet{gao2025reward}.

\paragraph{ImageNet 256$\times$256 Experiments.}
For high-resolution ImageNet experiments, we employ the SiT-XL/2 architecture as our backbone. We standardize the reward values to overcome the scale discrepancies in absolute values across different reward functions. The standardized formulation is defined as $\log w=\beta \frac{r-\mu}{\sigma+\epsilon}$, where $r$ denotes the raw reward, $\mu$ and $\sigma$ represent the exponential moving average (EMA) of the reward's mean and standard deviation, respectively, and $\epsilon$ is a small constant for numerical stability. We set the reward weighting parameter to $\beta = 1$ for the compressibility and incompressibility tasks, and $\beta = 3$ for CLIP-red and B\&W.

For our proposed method, the models are trained for a total of 16k iterations with a batch size of 32, utilizing the Adam optimizer \citep{kingma2014adam} ($\beta_1=0.5$, $\beta_2=0.9$) with a CosineAnnealingLR schedule and an initial learning rate of $4 \times 10^{-5}$ for the generator and $1.0 \times 10^{-4}$ for the discriminator. We apply an exponential moving average (EMA) to the generator weights with a decay rate of $0.9999$. The JKO step size is set to $h = 15$, with a default configuration of 20 steps $\times$ 800 inner iterations.

For the DPO baseline, we use a group size (batch size) of 8 and a total of 16k iterations, optimized via Adam \citep{kingma2014adam} with a learning rate of $1 \times 10^{-5}$ and gradient clipping at $1.0$. Exploration is performed over the time interval $[0.4, 0.7]$ with an online update interval of 40 iterations. To ensure a fair comparison and prevent batch skipping in sparse reward regimes, we use a reward-scale-dependent skip threshold, set to $0.05$ for standard tasks and lowered to $0.01$ for B\&W and CLIP-red, where the reward variance is narrower. FID is computed with 50k samples at 16k iterations.

\begin{figure}[t]
    \centering
    \setlength{\tabcolsep}{1pt}
    \begin{tabular}{cc}
        \includegraphics[width=0.48\linewidth]{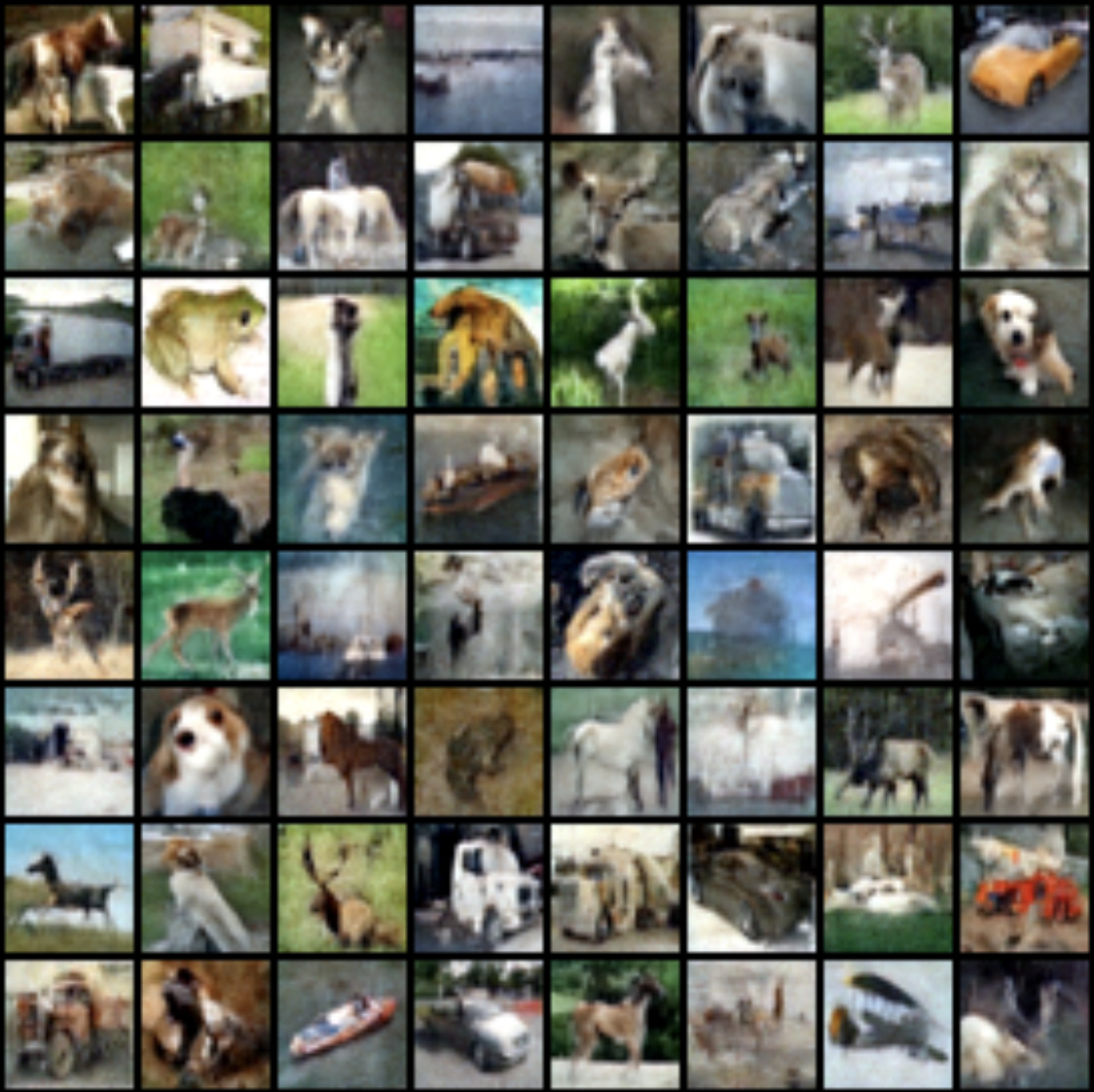} &
        \includegraphics[width=0.48\linewidth]{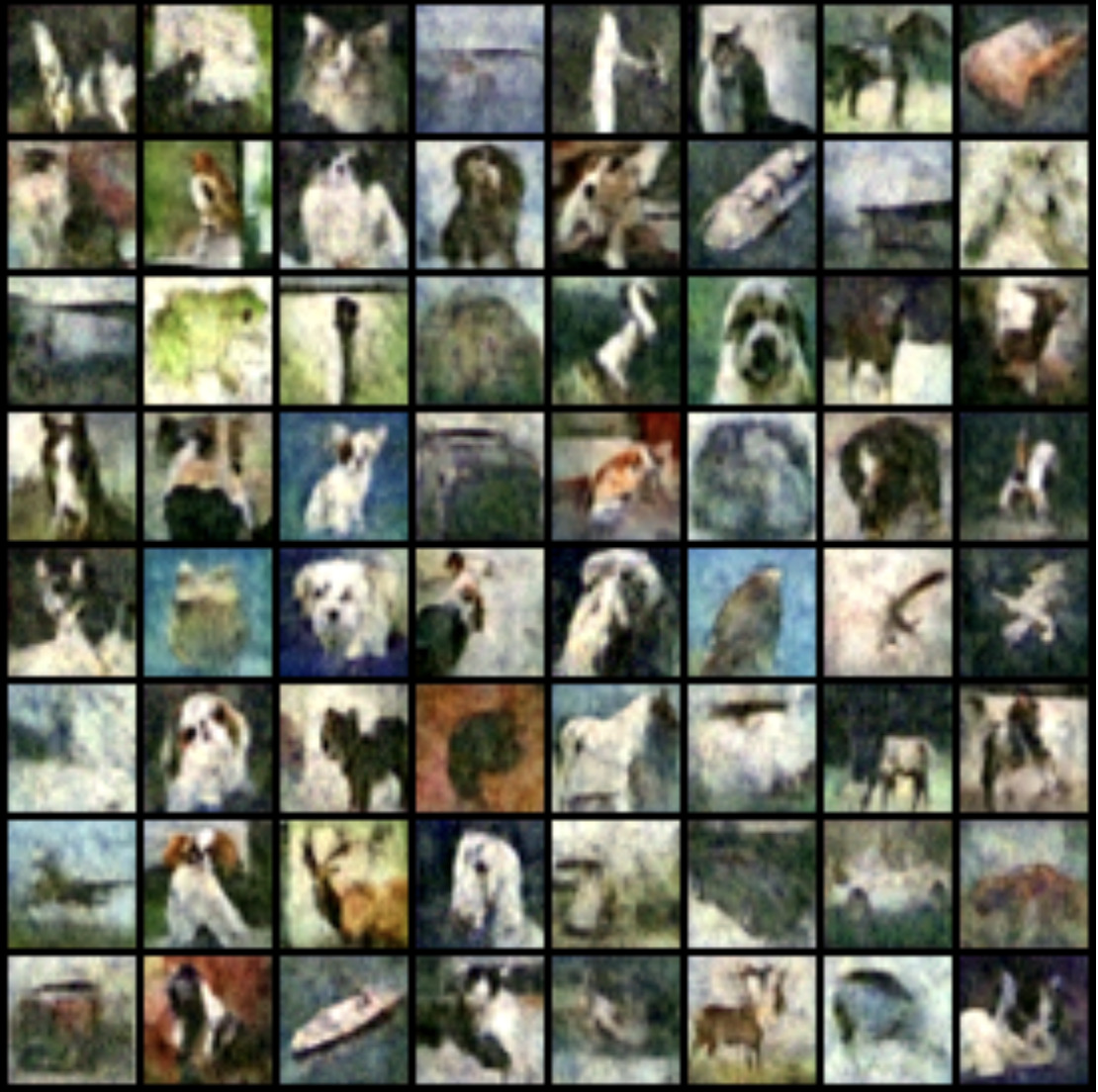} \\
        (a) Updated (800 iters) & (b) Fixed (800 iters) \\[2mm]
        \includegraphics[width=0.48\linewidth]{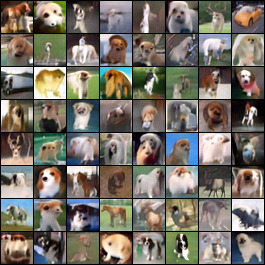} &
        \includegraphics[width=0.48\linewidth]{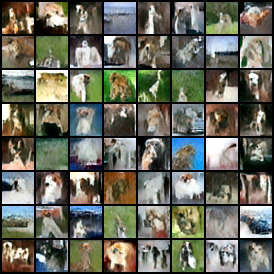} \\
        (c) Updated (5,000 iters) & (d) Fixed (5,000 iters) \\
    \end{tabular}
    \caption{\textbf{Qualitative comparison of proposal distributions on CIFAR-10.} At 800 iterations, both proposals remain close to the pre-trained distribution. At 5,000 iterations, \textit{Updated} produces clear dogs while preserving sharpness, whereas \textit{Fixed} produces washed-out samples with degraded structure.}
    \label{fig:ablation_proposal_samples}

\end{figure}

\section{Additional Results} \label{app:additional_results}

\subsection{Reward Curve on ImageNet 256$\times$256}
Figure~\ref{fig:imgnet_reward_curves} shows that 
our method achieves consistent reward improvement across all tasks throughout training. In contrast, DPO converges quickly in the early stage but saturates or even declines thereafter.

\subsection{Ablation Study on \textit{Fixed} vs. \textit{Updated} distribution.} \label{app:fixed_ablation}
In Figure~\ref{fig:ablation_proposal_samples}, we provide some qualitative comparisons for the ablation study on \textit{Fixed} vs. \textit{Updated} distribution described in the main paper. While \textit{Fixed} initially moves to the target distribution more quickly, it degrades into blurry samples as training proceeds, whereas \textit{Updated} retains the sample quality of the pretrained model.

\subsection{Additional Qualitative Results}

\begin{figure*}[htbp]
    \centering
    \begin{minipage}[c]{0.45\textwidth}
        \centering
        \textbf{(a) Pre-trained Model} \\[4pt]
        \includegraphics[width=1.0\linewidth]{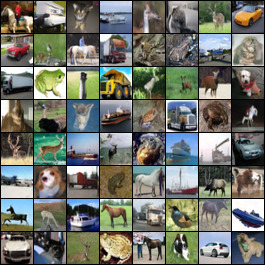}
    \end{minipage}
    \hfill
    \begin{minipage}[c]{0.45\textwidth}
        \centering
        \textbf{(b) Incompressibility} \\[4pt]
        \setlength{\tabcolsep}{2pt}
        \begin{tabular}{cc}
            \includegraphics[width=0.48\linewidth]{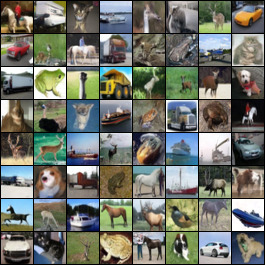} &
            \includegraphics[width=0.48\linewidth]{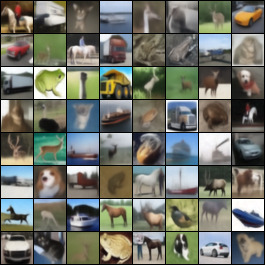} \\
            \small SFT & \small RWR \\
            \includegraphics[width=0.48\linewidth]{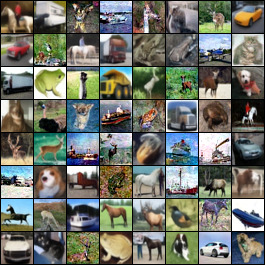} &
            \includegraphics[width=0.48\linewidth]{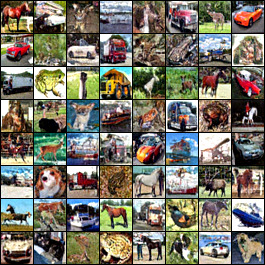} \\
            \small DPO & \small Ours
        \end{tabular}
    \end{minipage}

    \vspace{5pt}

    \begin{minipage}[c]{0.45\textwidth}
        \centering
        \textbf{(c) Compressibility} \\[4pt]
        \setlength{\tabcolsep}{2pt}
        \begin{tabular}{cc}
            \includegraphics[width=0.48\linewidth]{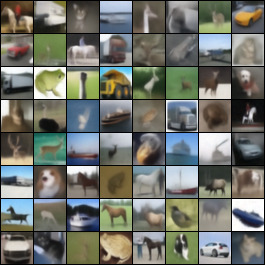} &
            \includegraphics[width=0.48\linewidth]{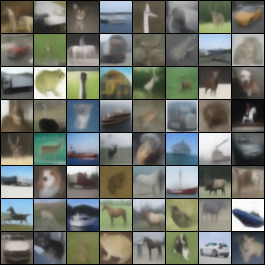} \\
            \small SFT & \small RWR \\
            \includegraphics[width=0.48\linewidth]{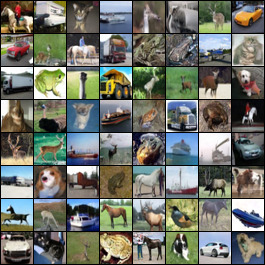} &
            \includegraphics[width=0.48\linewidth]{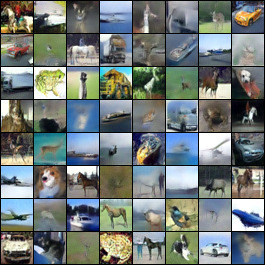} \\
            \small DPO & \small Ours
        \end{tabular}
    \end{minipage}
    \hfill
    \begin{minipage}[c]{0.45\textwidth}
        \centering
        \textbf{(d) Class 5 (Dog)} \\[4pt]
        \setlength{\tabcolsep}{2pt}
        \begin{tabular}{cc}
            \includegraphics[width=0.48\linewidth]{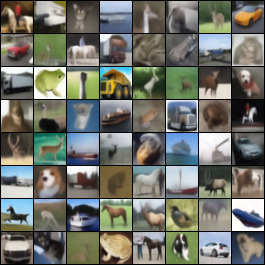} &
            \includegraphics[width=0.48\linewidth]{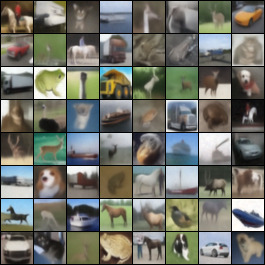} \\
            \small SFT & \small RWR \\
            \includegraphics[width=0.48\linewidth]{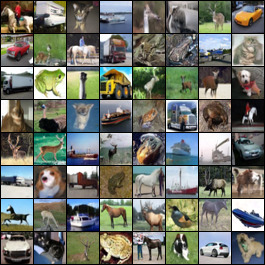} &
            \includegraphics[width=0.48\linewidth]{figures/app_samples/Ours_8x8.jpg} \\
            \small DPO & \small Ours
        \end{tabular}
    \end{minipage}

    \caption{
    \textbf{Qualitative comparisons across multiple reward tasks on CIFAR-10.} 
    (a) Pre-trained model samples generated from the initial noise. 
    (b) Incompressibility and (c) Compressibility represent non-differentiable reward optimization, while (d) Class 5 (Dog) represents differentiable reward optimization. 
    }
    \label{fig:cifar_full_grid}
\end{figure*}

\begin{figure*}[htbp]
    \centering
    \textbf{(a) Pre-trained Model} \\[4pt]
    \includegraphics[width=0.34\textwidth]{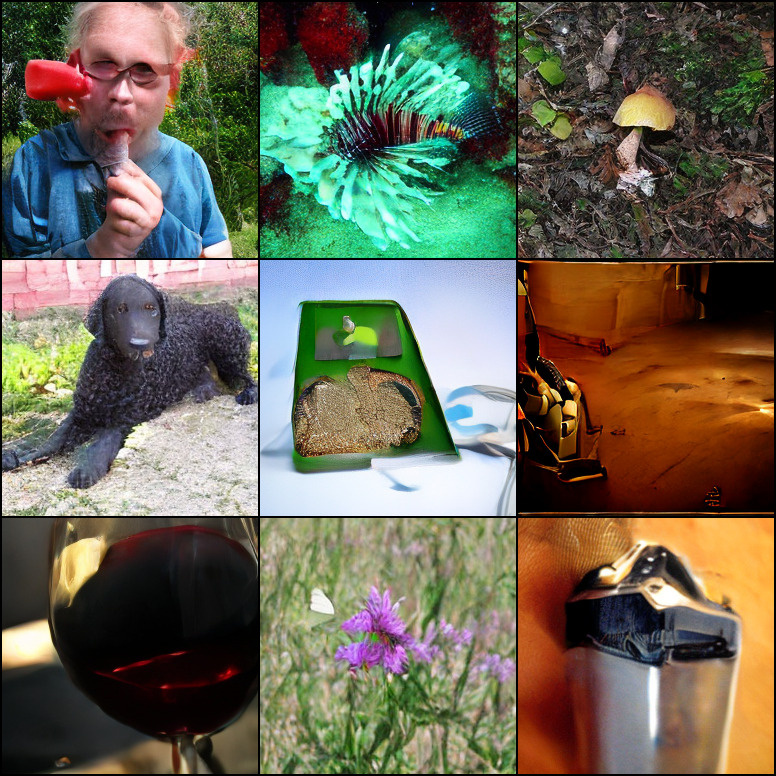}

    \vspace{14pt}

    \begin{minipage}[c]{0.49\textwidth}
        \centering
        \textbf{(b) Incompressibility} \\[4pt]
        \setlength{\tabcolsep}{1pt}
        \begin{tabular}{cc}
            \includegraphics[width=0.5\linewidth]{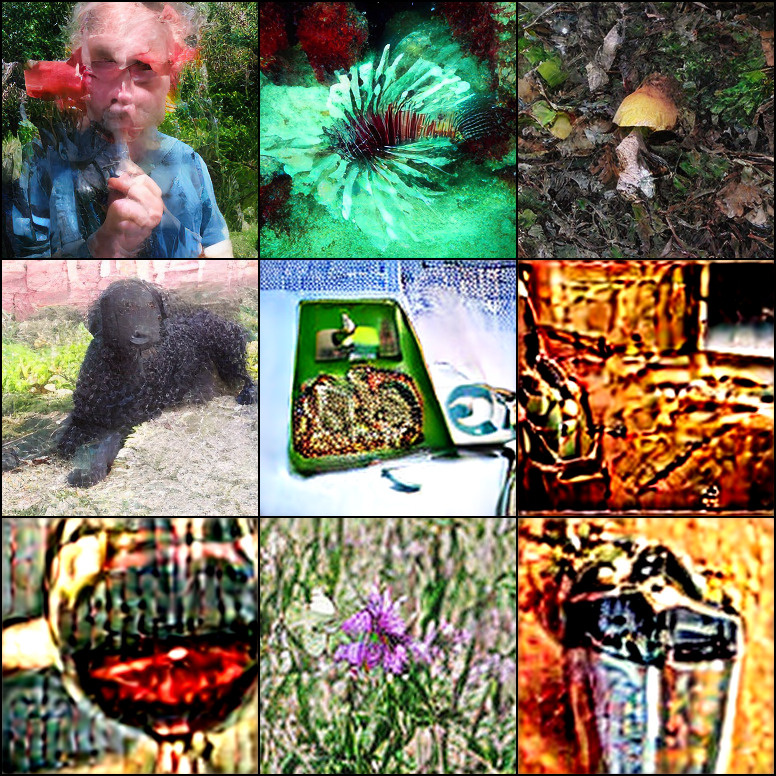} &
            \includegraphics[width=0.5\linewidth]{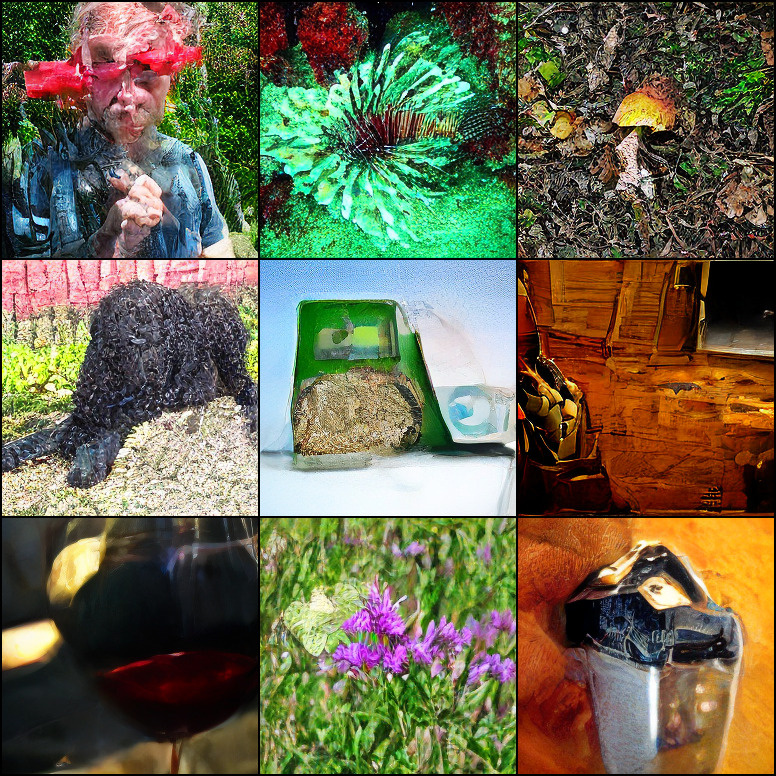} \\
            \small DPO & \small Ours
        \end{tabular}
    \end{minipage}
    \hfill
    \begin{minipage}[c]{0.49\textwidth}
        \centering
        \textbf{(c) Compressibility} \\[4pt]
        \setlength{\tabcolsep}{1pt}
        \begin{tabular}{cc}
            \includegraphics[width=0.5\linewidth]{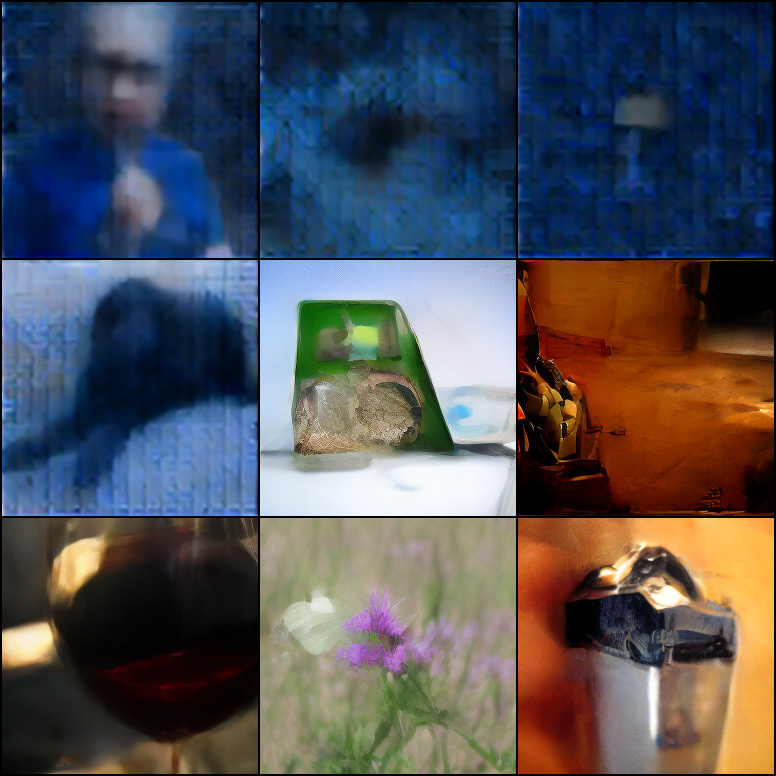} &
            \includegraphics[width=0.5\linewidth]{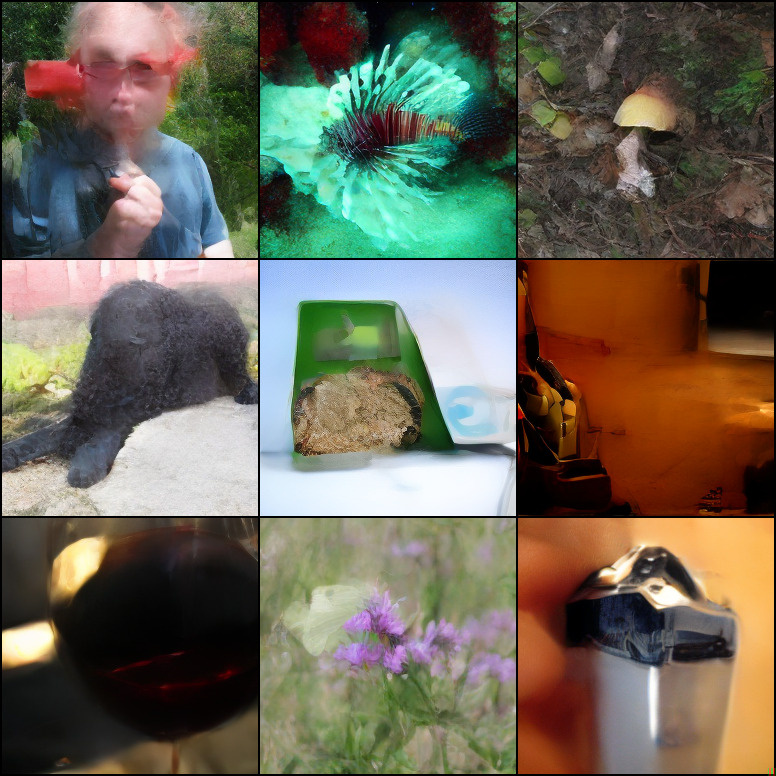} \\
            \small DPO & \small Ours
        \end{tabular}
    \end{minipage}

    \vspace{14pt}

    \begin{minipage}[c]{0.49\textwidth}
        \centering
        \textbf{(d) B\&W} \\[4pt]
        \setlength{\tabcolsep}{1pt}
        \begin{tabular}{cc}
            \includegraphics[width=0.5\linewidth]{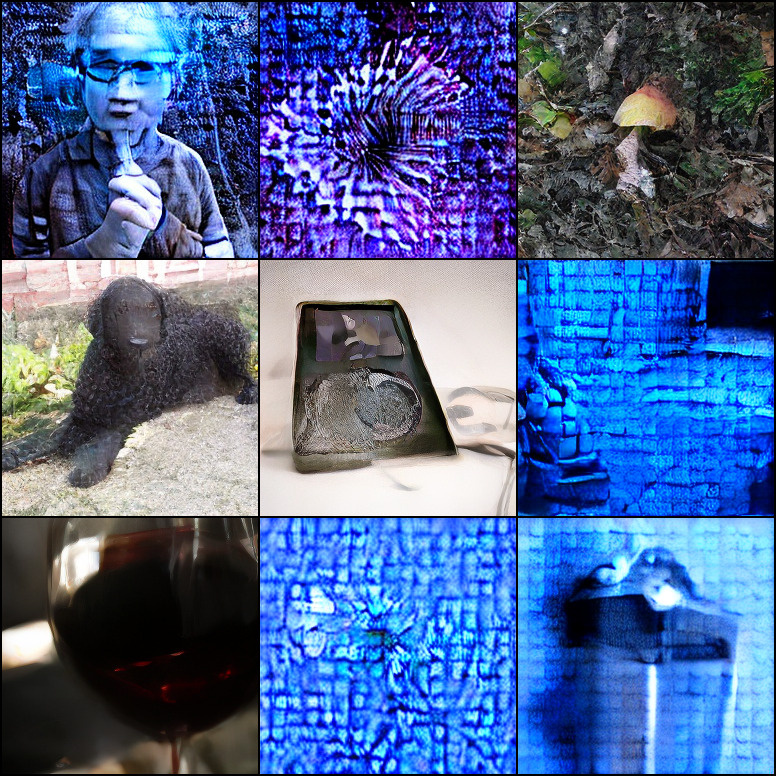} &
            \includegraphics[width=0.5\linewidth]{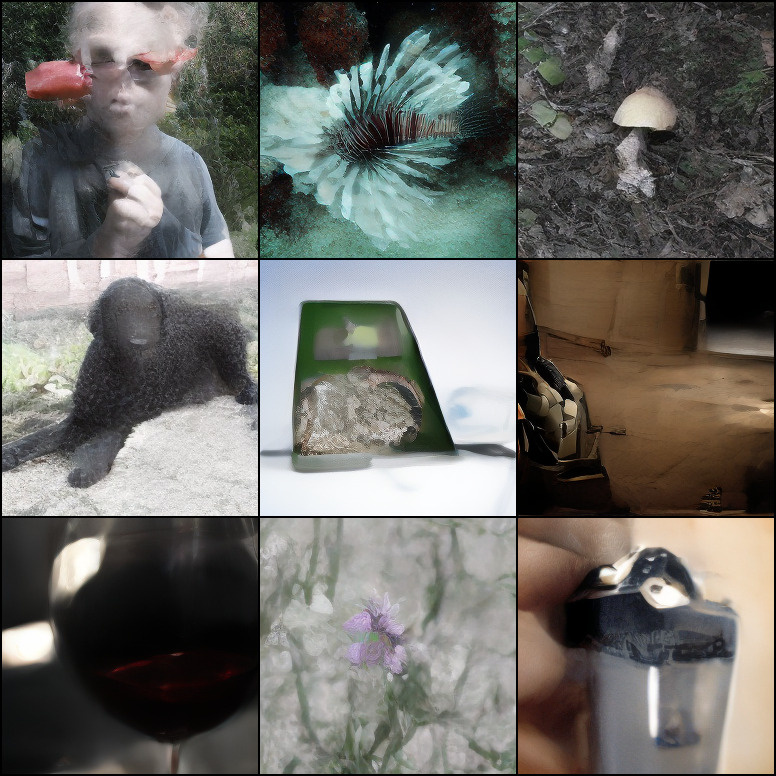} \\
            \small DPO & \small Ours
        \end{tabular}
    \end{minipage}
    \hfill
    \begin{minipage}[c]{0.49\textwidth}
        \centering
        \textbf{(e) CLIP-red} \\[4pt]
        \setlength{\tabcolsep}{1pt}
        \begin{tabular}{cc}
            \includegraphics[width=0.5\linewidth]{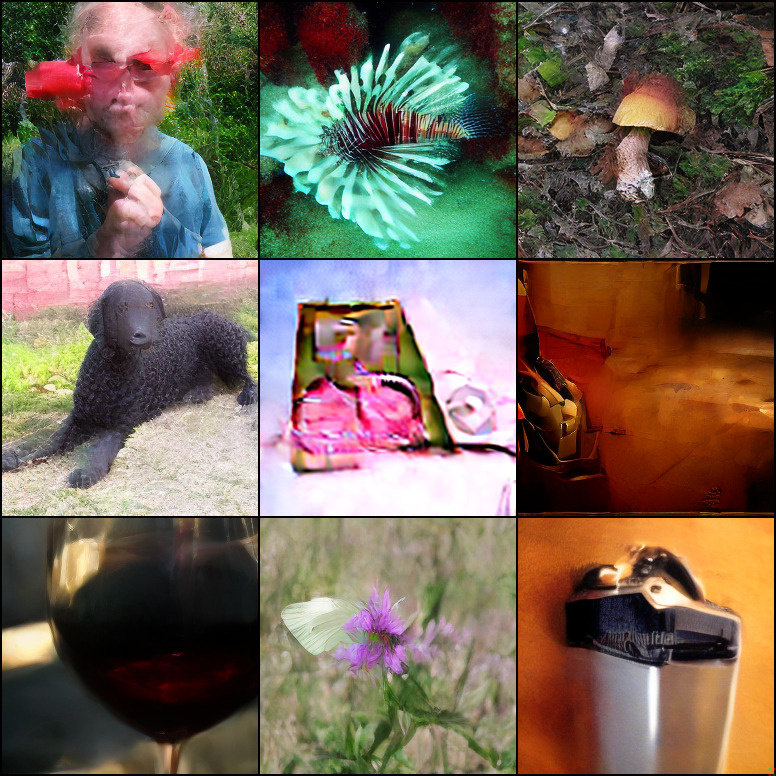} &
            \includegraphics[width=0.5\linewidth]{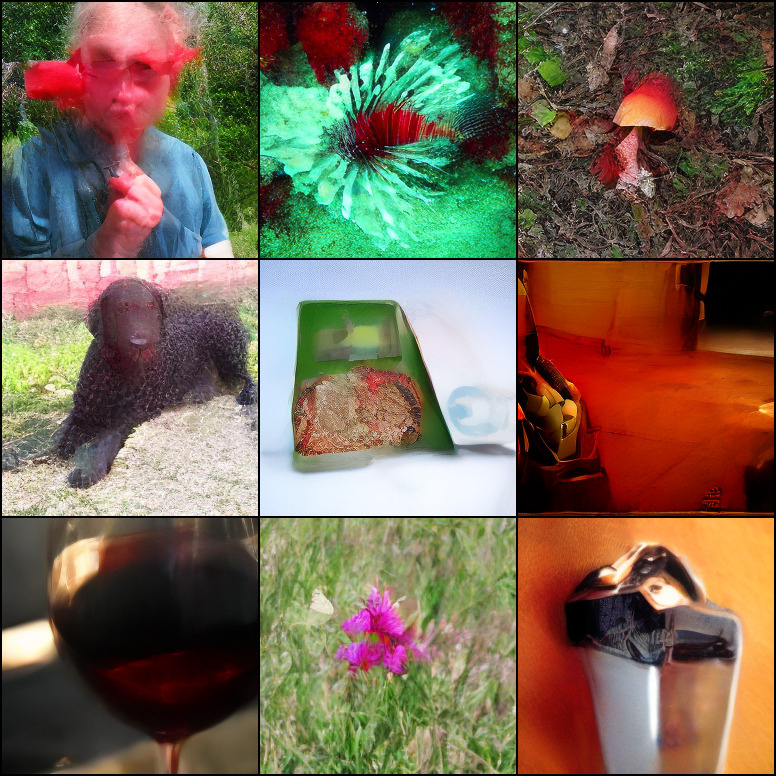} \\
            \small DPO & \small Ours
        \end{tabular}
    \end{minipage}
    \caption{
    \textbf{Qualitative comparisons across multiple reward tasks on ImageNet 256$\times$256.}
    (a) Pre-trained model samples generated from the initial noise, shown for reference.
    (b)--(e) show samples fine-tuned for each reward target under matched class labels and seeds.
    }
    \label{fig:imagenet_full_grid}
\end{figure*}

In Figure~\ref{fig:cifar_full_grid} and Figure~\ref{fig:imagenet_full_grid}, we present generated samples for our method and baselines on CIFAR-10 and ImageNet 256$\times$256.  

%% file: main.bbl
\begin{thebibliography}{41}
\providecommand{\natexlab}[1]{#1}

\bibitem[{Ambrosio, Gigli, and Savar{\'e}(2005)}]{ambrosio}
Ambrosio, L.; Gigli, N.; and Savar{\'e}, G. 2005.
\newblock \emph{Gradient flows: in metric spaces and in the space of probability measures}.
\newblock Springer Science \& Business Media.

\bibitem[{Black et~al.(2023)Black, Janner, Du, Kostrikov, and Levine}]{black2023training}
Black, K.; Janner, M.; Du, Y.; Kostrikov, I.; and Levine, S. 2023.
\newblock Training diffusion models with reinforcement learning.
\newblock \emph{arXiv preprint arXiv:2305.13301}.

\bibitem[{Chizat et~al.(2018)Chizat, Peyr{\'e}, Schmitzer, and Vialard}]{uot1}
Chizat, L.; Peyr{\'e}, G.; Schmitzer, B.; and Vialard, F.-X. 2018.
\newblock Unbalanced optimal transport: Dynamic and Kantorovich formulations.
\newblock \emph{Journal of Functional Analysis}, 274(11): 3090--3123.

\bibitem[{Choi, Choi, and Kang(2023)}]{uotm}
Choi, J.; Choi, J.; and Kang, M. 2023.
\newblock Generative Modeling through the Semi-dual Formulation of Unbalanced Optimal Transport.
\newblock In \emph{Thirty-seventh Conference on Neural Information Processing Systems}.

\bibitem[{Choi, Choi, and Kang(2024)}]{s-jko}
Choi, J.; Choi, J.; and Kang, M. 2024.
\newblock Scalable {W}asserstein Gradient Flow for Generative Modeling through Unbalanced Optimal Transport.
\newblock In \emph{Proceedings of the 41st International Conference on Machine Learning}, Proceedings of Machine Learning Research. PMLR.

\bibitem[{Clark et~al.(2024)Clark, Vicol, Swersky, and Fleet}]{clark2024directly}
Clark, K.; Vicol, P.; Swersky, K.; and Fleet, D. 2024.
\newblock Directly fine-tuning diffusion models on differentiable rewards.
\newblock In \emph{International Conference on Learning Representations}, volume 2024, 4793--4822.

\bibitem[{Deng et~al.(2009)Deng, Dong, Socher, Li, Li, and Fei-Fei}]{deng2009imagenet}
Deng, J.; Dong, W.; Socher, R.; Li, L.-J.; Li, K.; and Fei-Fei, L. 2009.
\newblock Imagenet: A large-scale hierarchical image database.
\newblock In \emph{2009 IEEE conference on computer vision and pattern recognition}, 248--255. Ieee.

\bibitem[{Fan et~al.(2023{\natexlab{a}})Fan, Liu, Ma, Zhou, and Chen}]{fanTMLR}
Fan, J.; Liu, S.; Ma, S.; Zhou, H.-M.; and Chen, Y. 2023{\natexlab{a}}.
\newblock Neural Monge Map estimation and its applications.
\newblock \emph{Transactions on Machine Learning Research}.

\bibitem[{Fan et~al.(2025)Fan, Shen, Cheng, Chen, Liang, and Liu}]{fan2025online}
Fan, J.; Shen, S.; Cheng, C.; Chen, Y.; Liang, C.; and Liu, G. 2025.
\newblock Online reward-weighted fine-tuning of flow matching with wasserstein regularization.
\newblock In \emph{The Thirteenth International Conference on Learning Representations}.

\bibitem[{Fan et~al.(2023{\natexlab{b}})Fan, Watkins, Du, Liu, Ryu, Boutilier, Abbeel, Ghavamzadeh, Lee, and Lee}]{fan2023dpok}
Fan, Y.; Watkins, O.; Du, Y.; Liu, H.; Ryu, M.; Boutilier, C.; Abbeel, P.; Ghavamzadeh, M.; Lee, K.; and Lee, K. 2023{\natexlab{b}}.
\newblock Dpok: Reinforcement learning for fine-tuning text-to-image diffusion models.
\newblock \emph{Advances in Neural Information Processing Systems}, 36: 79858--79885.

\bibitem[{Frans et~al.(2025)Frans, Hafner, Levine, and Abbeel}]{frans2025one}
Frans, K.; Hafner, D.; Levine, S.; and Abbeel, P. 2025.
\newblock One step diffusion via shortcut models.
\newblock In \emph{International Conference on Learning Representations}, volume 2025, 34668--34684.

\bibitem[{Gao, Zha, and Zhou(2025)}]{gao2025reward}
Gao, X.; Zha, J.; and Zhou, X.~Y. 2025.
\newblock Reward-directed score-based diffusion models via q-learning.
\newblock \emph{Journal of Machine Learning Research}, 26(302): 1--46.

\bibitem[{Geng et~al.(2025)Geng, Deng, Bai, Kolter, and He}]{geng2025mean}
Geng, Z.; Deng, M.; Bai, X.; Kolter, J.~Z.; and He, K. 2025.
\newblock Mean Flows for One-step Generative Modeling.
\newblock In \emph{Advances in Neural Information Processing Systems}.

\bibitem[{Gorbatovski et~al.(2025)Gorbatovski, Shaposhnikov, Malakhov, Surnachev, Aksenov, Maksimov, Balagansky, and Gavrilov}]{gorbatovski2025learn}
Gorbatovski, A.; Shaposhnikov, B.; Malakhov, A.; Surnachev, N.; Aksenov, Y.; Maksimov, I.; Balagansky, N.; and Gavrilov, D. 2025.
\newblock Learn your reference model for real good alignment.
\newblock In \emph{International Conference on Learning Representations}, volume 2025, 59414--59451.

\bibitem[{Heusel et~al.(2017)Heusel, Ramsauer, Unterthiner, Nessler, and Hochreiter}]{heusel2017gans}
Heusel, M.; Ramsauer, H.; Unterthiner, T.; Nessler, B.; and Hochreiter, S. 2017.
\newblock Gans trained by a two time-scale update rule converge to a local nash equilibrium.
\newblock \emph{Advances in neural information processing systems}, 30.

\bibitem[{Ho, Jain, and Abbeel(2020)}]{ho2020denoising}
Ho, J.; Jain, A.; and Abbeel, P. 2020.
\newblock Denoising Diffusion Probabilistic Models.
\newblock In \emph{Advances in Neural Information Processing Systems}, volume~33, 6840--6851.

\bibitem[{Jacq et~al.(2026)Jacq, Couairon, De~Bortoli, Berthet, Doucet, and Elie}]{jacq2026diffusion}
Jacq, A.; Couairon, G.; De~Bortoli, V.; Berthet, Q.; Doucet, A.; and Elie, R. 2026.
\newblock Diffusion Fine-tuning with Rewarded Moment Matching Distillation.
\newblock \emph{arXiv preprint arXiv:2606.30414}.

\bibitem[{Jia et~al.(2025)Jia, Nan, Zhao, and Liu}]{jia2025reward}
Jia, Z.; Nan, Y.; Zhao, H.; and Liu, G. 2025.
\newblock Reward Fine-Tuning Two-Step Diffusion Models via Learning Differentiable Latent-Space Surrogate Reward.
\newblock In \emph{Proceedings of the Computer Vision and Pattern Recognition Conference}, 12912--12922.

\bibitem[{Jordan, Kinderlehrer, and Otto(1998)}]{jko}
Jordan, R.; Kinderlehrer, D.; and Otto, F. 1998.
\newblock The variational formulation of the Fokker--Planck equation.
\newblock \emph{SIAM journal on mathematical analysis}, 29(1): 1--17.

\bibitem[{Kim et~al.(2024)Kim, Lai, Liao, Murata, Takida, Uesaka, He, Mitsufuji, and Ermon}]{kim2023consistency}
Kim, D.; Lai, C.-H.; Liao, W.-H.; Murata, N.; Takida, Y.; Uesaka, T.; He, Y.; Mitsufuji, Y.; and Ermon, S. 2024.
\newblock Consistency Trajectory Models: Learning Probability Flow ODE Trajectory of Diffusion.
\newblock In \emph{International Conference on Learning Representations}.

\bibitem[{Kingma and Ba(2014)}]{kingma2014adam}
Kingma, D.~P.; and Ba, J. 2014.
\newblock Adam: A method for stochastic optimization.
\newblock \emph{arXiv preprint arXiv:1412.6980}.

\bibitem[{Krizhevsky(2009)}]{Krizhevsky2009LearningML}
Krizhevsky, A. 2009.
\newblock Learning Multiple Layers of Features from Tiny Images.
\newblock Technical report, University of Toronto.

\bibitem[{Liero, Mielke, and Savar{\'e}(2018)}]{uot2}
Liero, M.; Mielke, A.; and Savar{\'e}, G. 2018.
\newblock Optimal entropy-transport problems and a new Hellinger--Kantorovich distance between positive measures.
\newblock \emph{Inventiones mathematicae}, 211(3): 969--1117.

\bibitem[{Lipman et~al.(2023)Lipman, Chen, Ben-Hamu, Nickel, and Le}]{lipman2023flow}
Lipman, Y.; Chen, R. T.~Q.; Ben-Hamu, H.; Nickel, M.; and Le, M. 2023.
\newblock Flow Matching for Generative Modeling.
\newblock In \emph{International Conference on Learning Representations (ICLR)}.

\bibitem[{Liu et~al.(2026)Liu, Choi, Chen, Miller, and Chen}]{asbs}
Liu, G.-H.; Choi, J.; Chen, Y.; Miller, B.~K.; and Chen, R.~T. 2026.
\newblock Adjoint schr{\"o}dinger bridge sampler.
\newblock \emph{Advances in Neural Information Processing Systems}, 38: 15673--15708.

\bibitem[{Liu et~al.(2025)Liu, Liu, Liang, Li, Liu, Wang, Wan, Zhang, and Ouyang}]{FlowGRPO}
Liu, J.; Liu, G.; Liang, J.; Li, Y.; Liu, J.; Wang, X.; Wan, P.; Zhang, D.; and Ouyang, W. 2025.
\newblock Flow-GRPO: Training Flow Matching Models via Online RL.
\newblock arXiv:2505.05470.

\bibitem[{Liu, Gong, and Liu(2023)}]{liu2022flow}
Liu, X.; Gong, C.; and Liu, Q. 2023.
\newblock Flow Straight and Fast: Learning to Generate and Transfer Data with Rectified Flow.
\newblock In \emph{International Conference on Learning Representations}.

\bibitem[{Meng et~al.(2023)Meng, Rombach, Gao, Kingma, Ermon, Ho, and Salimans}]{meng2023distillation}
Meng, C.; Rombach, R.; Gao, R.; Kingma, D.; Ermon, S.; Ho, J.; and Salimans, T. 2023.
\newblock On distillation of guided diffusion models.
\newblock In \emph{Proceedings of the IEEE/CVF conference on computer vision and pattern recognition}, 14297--14306.

\bibitem[{Oertell et~al.(2024)Oertell, Chang, Zhang, Brantley, and Sun}]{oertell2024rl}
Oertell, O.; Chang, J.~D.; Zhang, Y.; Brantley, K.; and Sun, W. 2024.
\newblock Rl for consistency models: Faster reward guided text-to-image generation.
\newblock \emph{arXiv preprint arXiv:2404.03673}.

\bibitem[{Peters and Schaal(2007)}]{peters2007reinforcement}
Peters, J.; and Schaal, S. 2007.
\newblock Reinforcement learning by reward-weighted regression for operational space control.
\newblock In \emph{Proceedings of the 24th international conference on Machine learning}, 745--750.

\bibitem[{Prabhudesai et~al.(2024)Prabhudesai, Goyal, Pathak, and Fragkiadaki}]{prabhudesai2023aligning}
Prabhudesai, M.; Goyal, A.; Pathak, D.; and Fragkiadaki, K. 2024.
\newblock Aligning Text-to-Image Diffusion Models with Reward Backpropagation.
\newblock arXiv:2310.03739.

\bibitem[{Radford et~al.(2021)Radford, Kim, Hallacy, Ramesh, Goh, Agarwal, Sastry, Askell, Mishkin, Clark et~al.}]{radford2021learning}
Radford, A.; Kim, J.~W.; Hallacy, C.; Ramesh, A.; Goh, G.; Agarwal, S.; Sastry, G.; Askell, A.; Mishkin, P.; Clark, J.; et~al. 2021.
\newblock Learning transferable visual models from natural language supervision.
\newblock In \emph{International conference on machine learning}, 8748--8763. PmLR.

\bibitem[{Rafailov et~al.(2023)Rafailov, Sharma, Mitchell, Manning, Ermon, and Finn}]{rafailov2023direct}
Rafailov, R.; Sharma, A.; Mitchell, E.; Manning, C.~D.; Ermon, S.; and Finn, C. 2023.
\newblock Direct preference optimization: Your language model is secretly a reward model.
\newblock \emph{Advances in neural information processing systems}, 36: 53728--53741.

\bibitem[{Rout, Korotin, and Burnaev(2022)}]{otm}
Rout, L.; Korotin, A.; and Burnaev, E. 2022.
\newblock Generative Modeling with Optimal Transport Maps.
\newblock In \emph{International Conference on Learning Representations}.

\bibitem[{Salimans and Ho(2022)}]{salimans2022progressive}
Salimans, T.; and Ho, J. 2022.
\newblock Progressive distillation for fast sampling of diffusion models.
\newblock \emph{arXiv preprint arXiv:2202.00512}.

\bibitem[{Schuhmann(2022)}]{schuhmann2022laion}
Schuhmann, C. 2022.
\newblock LAION Aesthetics predictor v1.

\bibitem[{Shekhar and Zhang(2025)}]{shekhar2025rocm}
Shekhar, S.; and Zhang, T. 2025.
\newblock ROCM: RLHF on consistency models.
\newblock \emph{arXiv preprint arXiv:2503.06171}.

\bibitem[{Shin et~al.(2026)Shin, Sul, Lee, Choi, and Choi}]{asbm}
Shin, J.; Sul, J.; Lee, J.; Choi, J.; and Choi, J. 2026.
\newblock Efficient Generative Modeling beyond Memoryless Diffusion via Adjoint Schr{\"o}dinger Bridge Matching.
\newblock \emph{arXiv preprint arXiv:2602.15396}.

\bibitem[{Song et~al.(2023)Song, Dhariwal, Chen, and Sutskever}]{song2023consistency}
Song, Y.; Dhariwal, P.; Chen, M.; and Sutskever, I. 2023.
\newblock Consistency Models.
\newblock In \emph{International Conference on Machine Learning}.

\bibitem[{Song et~al.(2021)Song, Sohl-Dickstein, Kingma, Kumar, Ermon, and Poole}]{song2021scorebased}
Song, Y.; Sohl-Dickstein, J.; Kingma, D.~P.; Kumar, A.; Ermon, S.; and Poole, B. 2021.
\newblock Score-Based Generative Modeling through Stochastic Differential Equations.
\newblock In \emph{International Conference on Learning Representations}.

\bibitem[{Villani et~al.(2009)}]{villani}
Villani, C.; et~al. 2009.
\newblock \emph{Optimal transport: old and new}, volume 338.
\newblock Springer.

\end{thebibliography}
